\documentclass[runningheads]{llncs}

\usepackage{eccv}

\usepackage{eccvabbrv}

\usepackage{graphicx}
\usepackage{booktabs}

\usepackage{pgfplots}

\usepackage{multirow}
\usepackage{booktabs}

\usepackage{wrapfig}

\usepackage[pagebackref,breaklinks,colorlinks,citecolor=eccvblue]{hyperref}

\begin{document}
	
	\title{Self-supervised Dataset Distillation: A Good {\em{Compression}} Is All You Need} 
	
	\titlerunning{Self-supervised Dataset Distillation}
	
	\author{Muxin Zhou \and
		Zeyuan Yin \and
		Shitong Shao \and
		Zhiqiang Shen\thanks{Corresponding author.}}
	
	\authorrunning{M.~Zhou et al.}
	
	\institute{VILA Lab, Mohamed bin Zayed University of AI \\
		\email{\{muxin.zhou,zeyuan.yin,zhiqiang.shen\}@mbzuai.ac.ae, 1090784053sst@gmail.com}\\
		\url{https://github.com/VILA-Lab/SRe2L/tree/main/SCDD/} 
	}
	
	\maketitle

	\begin{center}
		\centering
		\includegraphics[width=0.98\linewidth]{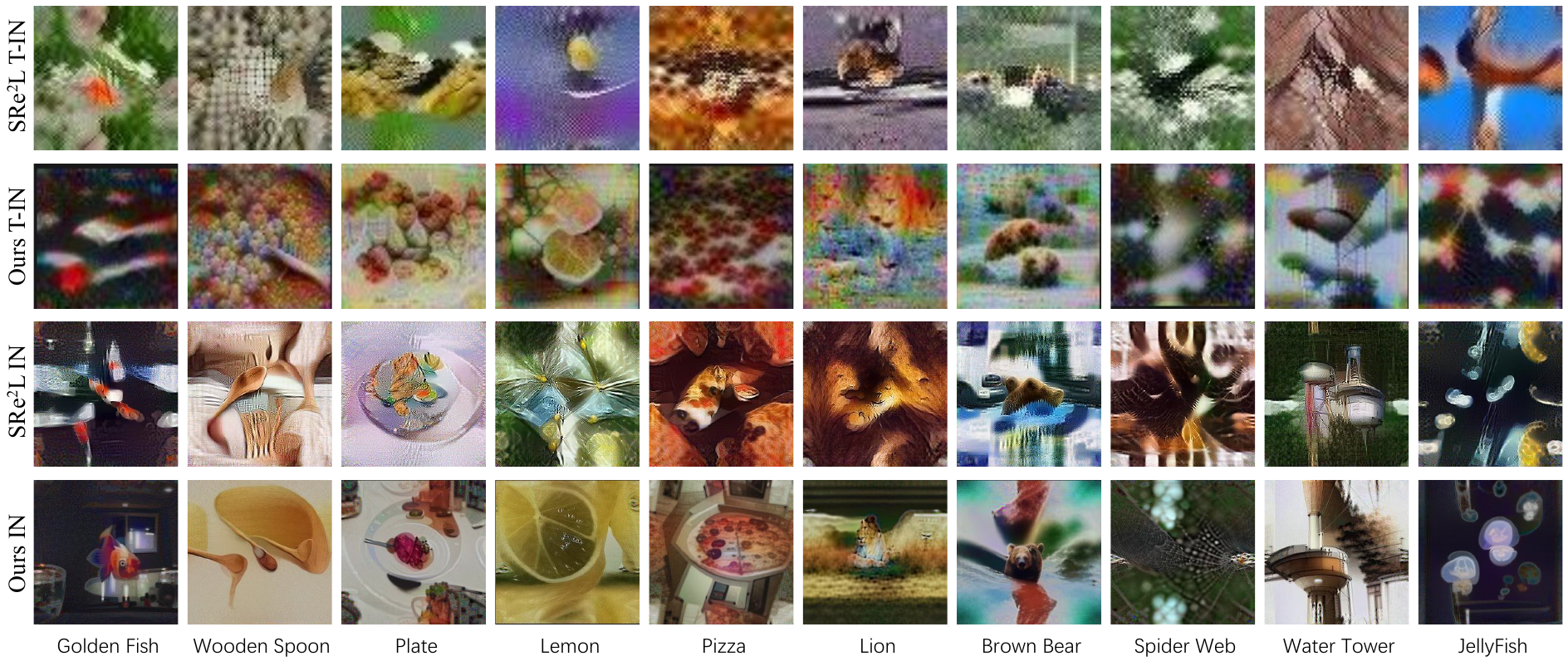}
		\vspace{-0.15in}
		\captionof{figure}{Example distilled images from SRe$^2$L~\cite{yin2023squeeze} and our 64$\times$64 Tiny-ImageNet (top two rows), 224$\times$224 ImageNet-1K (bottom two rows). All our synthetic data is generated from the self-supervised pretrained models, while the more realistic images with better semantic alignment and details are obtained. Moreover, training a conventional deep model with our distilled images results in a model that achieves test accuracy on the original validation data markedly superior to previous dataset distillation methods. {More visualization results are available at \href{https://drive.google.com/file/d/1uQgGPx36WkBH-qh6iTpx2H2Dn86qXRAR/view?usp=sharing}{link}.}}
		\label{fig:vis_all}
	\end{center}%

	\begin{abstract}
		Dataset distillation aims to {{compress}} information from a large-scale original dataset to a new compact dataset while striving to preserve the utmost degree of the original data informational essence. Previous studies have predominantly concentrated on aligning the intermediate statistics between the original and distilled data, such as weight trajectory, features, gradient, BatchNorm, etc. 
		In this work, we consider addressing this task through the new lens of {\bf model informativeness} in the compression stage on the original dataset pretraining. We observe that with the prior state-of-the-art SRe$^2$L, as model sizes increase, it becomes increasingly challenging for supervised pretrained models to recover learned information during data synthesis, as the channel-wise mean and variance inside the model are flatting and less informative. We further notice that larger variances in BN statistics from self-supervised models enable larger loss signals to update the recovered data by gradients, enjoying more informativeness during synthesis. Building on this observation, we introduce SC-DD, a simple yet effective {\bf  S}elf-supervised {\bf  C}ompression framework for {\bf  D}ataset {\bf  D}istillation that facilitates diverse information compression and recovery compared to traditional supervised learning schemes, further reaps the potential of large pretrained models with enhanced capabilities. 
		Extensive experiments are conducted on CIFAR-100, Tiny-ImageNet and ImageNet-1K datasets to demonstrate the superiority of our proposed approach. The proposed SC-DD outperforms all previous state-of-the-art supervised dataset distillation methods when employing larger models, such as SRe$^2$L, MTT, TESLA, DC, CAFE, etc., by large margins under the same recovery and post-training budgets. 
	\end{abstract}
	
	\section{Introduction}
	\label{sec:intro}
	
\begin{wrapfigure}{l}{0.5\textwidth}
		\centering
		\vspace{-0.26in}
		\includegraphics[width=0.999\linewidth]{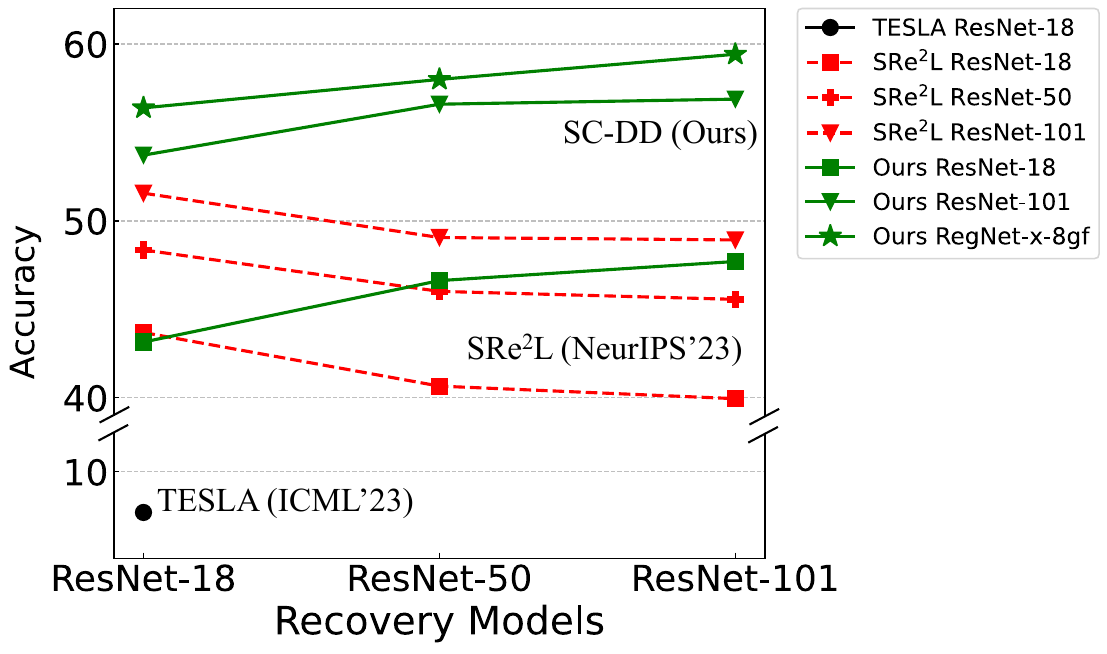}
		\vspace{-0.25in}
		\caption{Top-1 accuracy of SRe$^2$L~\cite{yin2023squeeze} and our approach on full ImageNet-1K with recovery model scales from small to large. The recovery budget is 1$k$ iterations. Each curve presents the post validation on ResNet-\{18, 50, 101\} and RegNet-x-8gf.} 
		\label{fig:compare}
		\vspace{-0.3in}
	\end{wrapfigure}

	Large-scale datasets and models are two major thrusts for the current remarkable achievements in the domains of computer vision~\cite{krizhevsky2012imagenet,he2016deep,dosovitskiy2021image}, natural language processing~\cite{kenton2019bert,brown2020language,openai2023gpt4} and speech~\cite{gulati2020conformer,hsu2021hubert}. 
	In the field of dataset distillation, several studies, including MTT~\cite{cazenavette2022dataset}, SRe$^2$L~\cite{yin2023squeeze}, TESLA~\cite{cui2023scaling}, and CDA~\cite{yin2023dataset}, have emphasized the crucial role of large-scale datasets. Yet, the significance of large-scale models within dataset distillation task remains underacknowledged. The recent SRe$^2$L approach~\cite{yin2023squeeze} suggests using a more expansive squeezing model for recovery. However, as these model sizes expand, supervised pretrained models in this approach face growing difficulties in retrieving learned knowledge during data synthesis with inferior performance. As shown in Fig.~\ref{fig:compare}, SRe$^2$L experiences significant declines in performance as the size of the recovery models increases.
	
	\begin{figure*}[t]
		\centering
		\includegraphics[width=0.86\linewidth]{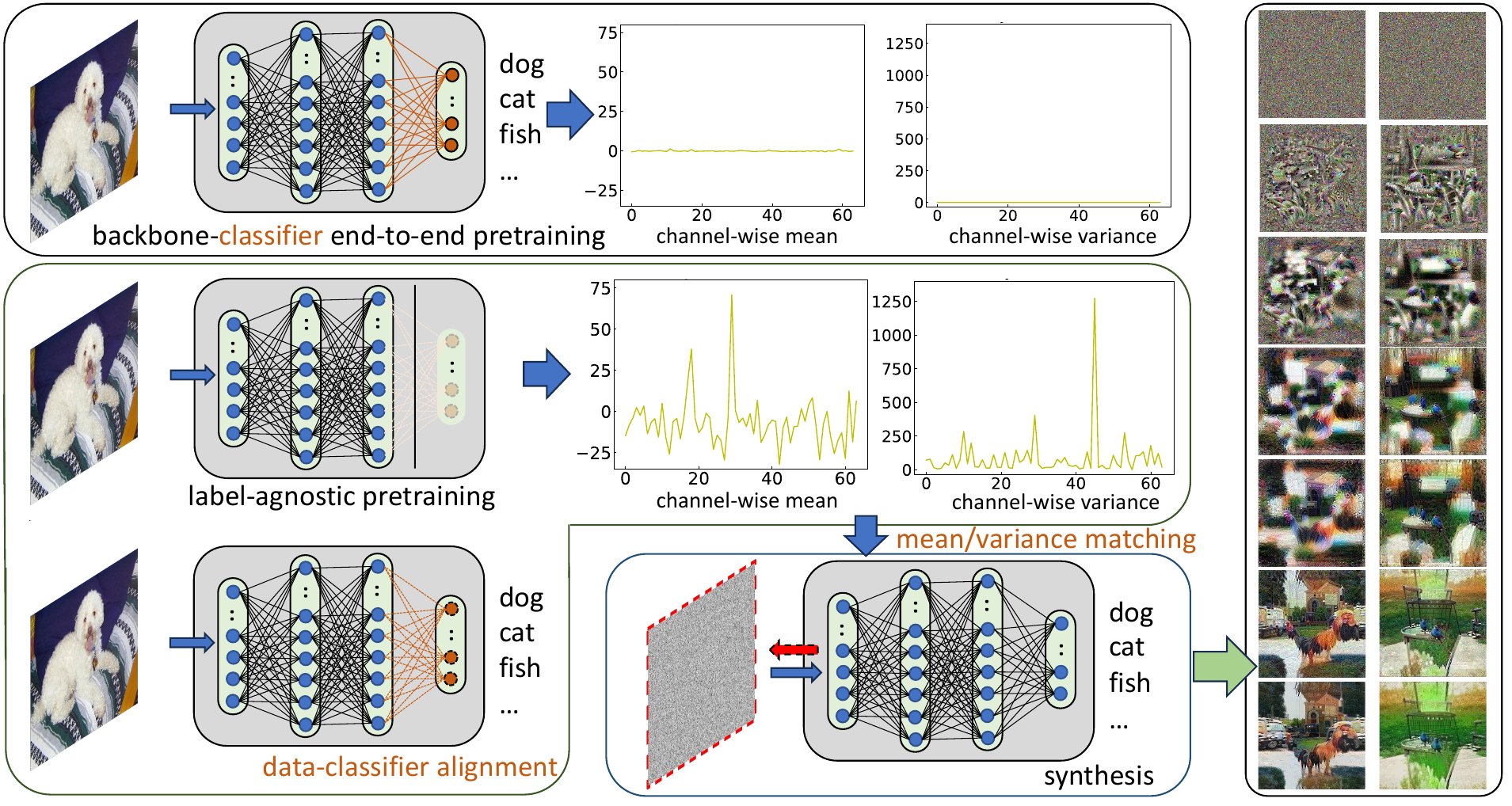}
		\vspace{-0.1in}
		\caption{Overview of our learning paradigm. The top-left subfigure is the paradigm of supervised pertaining with an end-to-end training scheme for both the backbone network and final alignment classifier. The bottom-left subfigure is the paradigm of our proposed procedure for dataset distillation: a backbone model is first pretrained using a self-supervised objective, then a linear probing layer is adjusted to align the distribution of pertaining and target dataset distribution. We do not fine-tune the backbone during the alignment phase to preserve the better intermediate distributions of mean and variance in batch normalization layers (illustrated in the middle yellow line chart of the figure). The bottom-middle subfigure is the data synthesis procedure and the left subfigure is the visualization of distilled images.}
		\label{overview}
		\vspace{-0.1in}
	\end{figure*}
	
	In this work, we tackle the challenges associated with the complexities of {\em compression} during the pertaining stage. We investigate the reasons behind the supervised learned model's inefficacy in dataset distillation, especially as the model size increases. Our observations reveal that when a model undergoes supervised pretraining, the BN statistic distributions of both intermediate mean and variance across channels in a layer trend towards uniformity, as illustrated in Fig.~\ref{overview} (the top-right two sub-figures\footnote{x-axis represents the channel index and y-axis represents the value.}). Moreover, the larger the model, the more even and flatting this statistic distribution appears. Given that SRe$^2$L~\cite{yin2023squeeze} employs these distributions as layer-wise and channel-wise labels to retrieve pretrained dataset details, the flattened distributions make it challenging for the model to discern and retain the most important and fine-grained information from the intricate knowledge.
	
	To mitigate the limitations of inadequate supervisory signals in data synthesis from the current pretraining scheme, we initiate our investigation by examining how different pretraining objectives affect the distribution of intermediate features. This analysis leads to a basic understanding of why existing advanced dataset distillation approaches experience a decline in performance as the size of model increases\footnote{Here, we indicate the model size in recovery stage instead of post-validation stage.}. We then present a distributional pertaining paradigm that focuses on amplifying the magnitude of intermediate mean and variance of the backbone, in line with a model-label semantic alignment learning process. This resembles the traditional self-supervised pretraining combined with the linear probing method. However, it is important to note that our pretraining does not always align with the same datasets. For instance, we might pretrain our backbone using the larger-scale ImageNet-21K and subsequently align it with ImageNet-1K dataset, leveraging the richer information in the larger dataset. 
	
	We present a concrete definition of {\em compression} in our framework, and explore its significance in the context of dataset distillation. 
	The ability to {\em compress} a dataset by a model is partially measured by the accuracy of the model derived. In the scenario of supervised training, this pertains directly to the accuracy achieved on validation data. When dealing with a self-supervised learning approach, we employ linear probing, solely tuning the final classifier to align with the semantic labels. However, a good {\em compression} method in dataset distillation means the learned model can recover more informative data which is beneficial in subsequent training on synthetic data. We observe that a self-supervised learned model with linear probing (to preserve distribution on the backbone learned during compression) will have lower accuracy in the pretraining stage, but is easier for data synthesis and post-training in dataset distillation task. 
	
	Our framework enjoys at least the following three advantages: {\bf (1)} Amplified supervision signals in data synthesis from the pertaining. {\bf (2)} A positive correlation between model size in recovery stage and performance in post-training stage, which provides more potential for scaling-up this problem. {\bf (3)} The simplicity of whole learning process while offering state-of-the-art performance. We highlight that {\em large models matter for data synthesis} in dataset distillation, but previous approaches failed to employ large-scale models in both data synthesis and post-training phases on this task. This work represents the inaugural and pioneering effort to integrate self-supervised pretraining into the dataset distillation, forming a simple approach that currently yields competitive accuracy. 
	The proposed procedure can be regarded as a decoupling process of intermediate feature distribution learning and high-level semantic alignment, which entirely differs from previous dataset distillation solutions. Our approach consistently improves the performance when the size of recovering model becomes larger.
	
	Despite its conceptual simplicity, we show empirically that with the proposed solution, namely {\bf S}elf-supervised {\bf  C}ompression for {\bf  D}ataset {\bf  D}istillation (SC-DD), our approach achieves significantly better accuracy than the prior state-of-the-art MTT, SRe$^2$L, TESAL, DM with the better scalability to larger recovery model architectures. The performance gains are evident across various datasets: SC-DD obtains 53.4\% on CIFAR-100, 45.9\% on Tiny-ImageNet, and 53.1\% on ImageNet-1K under IPC 50, surpassing previous the best by remarkable margins. More importantly, our approach demonstrates a positive correlation between model size and performance, that is, as the recovery model scales, we observe a corresponding uptick in efficacy, thus amplifying its prospective advantages. 
	This characteristic gains heightened significance in the era dominated by large-scale models and datasets, providing indispensable value for scaling up dataset distillation task.
	
	Our contributions:
	\vspace{-0.05in}
	\begin{itemize}
		\addtolength{\itemsep}{-0.00in}
		\item {We identify the drawback of previous state-of-the-art approach in dataset distillation that larger recovering models are consistently inferior under the supervised learning scheme. We uncover that the reason stems from the less informative mean and variance distributions for data synthesis.}
		\item {To our best knowledge, this is the pioneering work to reveal that self-supervised intermediate distributions are more informative for dataset distillation. Our work highlights the importance of pertaining scheme, and we provide detailed intuitions and analyses both empirically and theoretically.}
		\item {Extensive experiments are conducted on various CIFAR-100, Tiny-ImageNet, ImageNet-1K datasets and diverse model architectures. The proposed self-supervised learning framework outperforms all prior supervised dataset distillation counterparts by significant margins.}
	\end{itemize}
	
	\section{Approach}
	
	\vspace{-0.05in}
	\vspace{-0.04in}
	\noindent{\bf Preliminaries: Dataset Distillation.} Given an original large-scale dataset, Dataset Distillation (DD) seeks to generate a significantly condensed dataset comprised of synthetic samples with corresponding one-hot or soft labels. Despite its smaller size, models trained on this distilled dataset can achieve performance levels akin to those trained on the original dataset. 
	Let $\mathcal{D}_l=\left\{\left(\boldsymbol{x}_{i}, \boldsymbol{y}_{i}\right)\right\}_{i=1}^{|\mathcal{D}_l|}$ be a large labeled dataset, our goal is to synthesize a more concise distilled dataset, which we will denote as 
	$\mathcal{D}_{{d}}=\left\{\left(\boldsymbol{x}_i^{\prime},\boldsymbol{y}_i^{\prime}\right)\right\}_{i=1}^{|\mathcal{D}_d|}$. 
	In this distilled dataset, $\boldsymbol{y}^{\prime}$ represents either a hard or soft label corresponding to the synthetic data $\boldsymbol{x}^{\prime}$. It is also worth noting that the size of the distilled dataset, $\left|\mathcal{D}_d\right|$, is significantly smaller than the original dataset $\left|\mathcal{D}_l\right|$. Nonetheless, $\left|\mathcal{D}_d\right|$ retains the crucial information from $\left|\mathcal{D}_l\right|$. Our learning task is then defined on this distilled synthetic dataset: 
	\begin{equation}
		\begin{gathered}
			\boldsymbol{\theta}_{\mathcal{D}_d}=\underset{\boldsymbol{\theta}}{\arg \min } \mathcal{L}_{\mathcal{D}_d}(\boldsymbol{\theta}) 
		\end{gathered}
	\end{equation}
	where $\theta$ is the model weights in post-training on synthetic data. $\mathcal{L}$ is the objective function, e.g., cross-entropy loss.
	\begin{equation}
		\begin{gathered}
			\mathcal{L}_{\mathcal{D}_d}(\boldsymbol{\theta})=\mathbb{E}_{\left(\boldsymbol{x}^{\prime}, \boldsymbol{y}^{\prime}\right) \in \mathcal{D}_d}\left[\ell\left(\mathcal{M}_{\boldsymbol{\theta}_{\mathcal{D}_d}}\left(\boldsymbol{x}^{\prime}\right), \boldsymbol{y}^{\prime}\right)\right]
		\end{gathered}
	\end{equation}
	where $\mathcal{M}_{\boldsymbol{\theta}_{\mathcal{D}_d}}$ is the target model to train. The concrete $\ell$ formulation can be the soft cross-entropy loss if $\boldsymbol{y}^{\prime}$ is a soft label. The goal of dataset distillation is to generate synthetic data that aspires to achieve either a specific or minimal deviation in performance on the original dataset. This comparison is drawn between models trained on the synthetic data and that trained on the original dataset. Consequently, our goal is to optimize the synthetic dataset $\mathcal{D}_d$ accordingly:
	\vspace{-0.04in}
	\begin{equation}
		\begin{gathered}
			\underset{\mathcal{D}_d,|\mathcal{D}_d|}{\arg \min }(\sup \{|\ell(\mathcal{M}_{\boldsymbol{\theta}_{\mathcal{D}_l}}(\boldsymbol{x}_{\mathrm{val}}), \boldsymbol{y}_{\mathrm{val}})  - \ell(\mathcal{M}_{\boldsymbol{\theta}_{\mathcal{D}_d}}(\boldsymbol{x}_{\mathrm{val}}), \boldsymbol{y}_{\mathrm{val}})|\}_{(\boldsymbol{x}_{\mathrm{val}}, \boldsymbol{y}_{\mathrm{val}}) \sim \mathcal{D}_l})
		\end{gathered}
	\end{equation}
	where $\boldsymbol{x}_{\mathrm{val}}$ and $\boldsymbol{y}_{\mathrm{val}}$ are the real validation data and corresponding label in the original dataset $\mathcal{D}_l$. 
	In the procedure, we learn both $<$\texttt{data}, \texttt{label}$>\in\!\mathcal{D}_d$ with the corresponding number of distilled data in each class.
	
	\noindent{\bf Previous Solutions on Large-scale Datasets.} SRe$^2$L~\cite{yin2023squeeze} represents the first approach to effectively compress the large-scale ImageNet-1K dataset while preserving its vital information and performance attributes. This method unfolds across three distinct phases: initially, a model undergoes training from scratch in a supervised scheme, ensuring it captures the majority of the significant information from the original dataset. Subsequently, in the second phase, a recovery procedure is employed to generate the intended data from Gaussian noise. Finally, in the third phase, the generated synthetic data undergoes a crop-level relabeling process to accurately represent the synthetic data's actual soft labels.
	
	\vspace{-0.1in}
	\subsection{Understanding Dataset Compression}
	\vspace{-0.04in}
	
	Typically, we can employ supervised learning (SL) or self-supervised learning (SSL) paradigms to compress knowledge from original large-scale datasets like ImageNet-1K and store information into a trained model with dense parameters for the subsequent synthesizing of distilled data. Fig.~\ref{overview} illustrates an overview of our learning paradigm. Recently, self-supervised representation learning approaches have demonstrated superior performance on representation capability over supervised models, especially in downstream tasks such as image recognition and dense prediction. In this section, we first discuss the expressivity and generalization of self-supervised pertaining for dataset distillation task, and then analyze the essence that the self-supervised representations of the mean and variance statistics are more informative than the supervised models.
	
	\noindent{\bf Expressivity}. This refers to the ability of a model to capture a wide variety of underlying patterns and structures in original data. We aim to address ``{\em Can the representation learned from SSL accurately express the inherent distributions for categories?}'' Prior study \cite{lee2021predicting} has proven that training a linear layer upon SSL yields a small approximation error for complex ground truth function class and can substantially reduce labeled sample complexity, which indicates SSL models are more expressive than supervised ones. Our visualization in Fig.~\ref{fig:vis_all} also supports this statement.
	
	\noindent{\bf Generalization}. 
	We aim to ensure that the data generated from various pretraining approaches achieves enhanced generalization across diverse architectures and tasks, akin to the performance observed on the real datasets. The prior study \cite{huang2023towards} in SSL presented a measurement to offer an upper bound for the generalization ability in downstream classification tasks. This approach with the measure uncovers that the effectiveness of contrastive self-supervised learning hinges on three critical factors of the alignment of positive samples, the divergence of class centers, and the density of augmented data. The alignment pertains to the characteristics of the better learned representations by SSL.
	
	\noindent{\bf Our Observations}. Distinct objective functions such as SL and SSL invariably result in varying feature distributions. It has been noted that a supervised loss stabilizes the intermediate distribution, whereas a self-supervised loss emphasizes making the features more distinct. Furthermore, models trained with SSL demonstrate greater ease in synthesizing data during the process of dataset distillation. In this work, we focus on data synthesis that leverages pretrained representation distributions. We parameterize the final predictor as follows: given features $f(x) \! \in \! \mathbb{R}^k$ for some feature extractor parameters $w \! \in \! \mathcal{W}$, and a linear ``head'' $v \! \in \! \mathcal{V}$, we have $f_{v, w}(x)\!=\!v^{\top} f(x)$. In our experiments, $f$ is a deep network and $v$ is a linear projection.
	
	\begin{figure*}[h]
		\centering
		\includegraphics[width=0.9\linewidth]{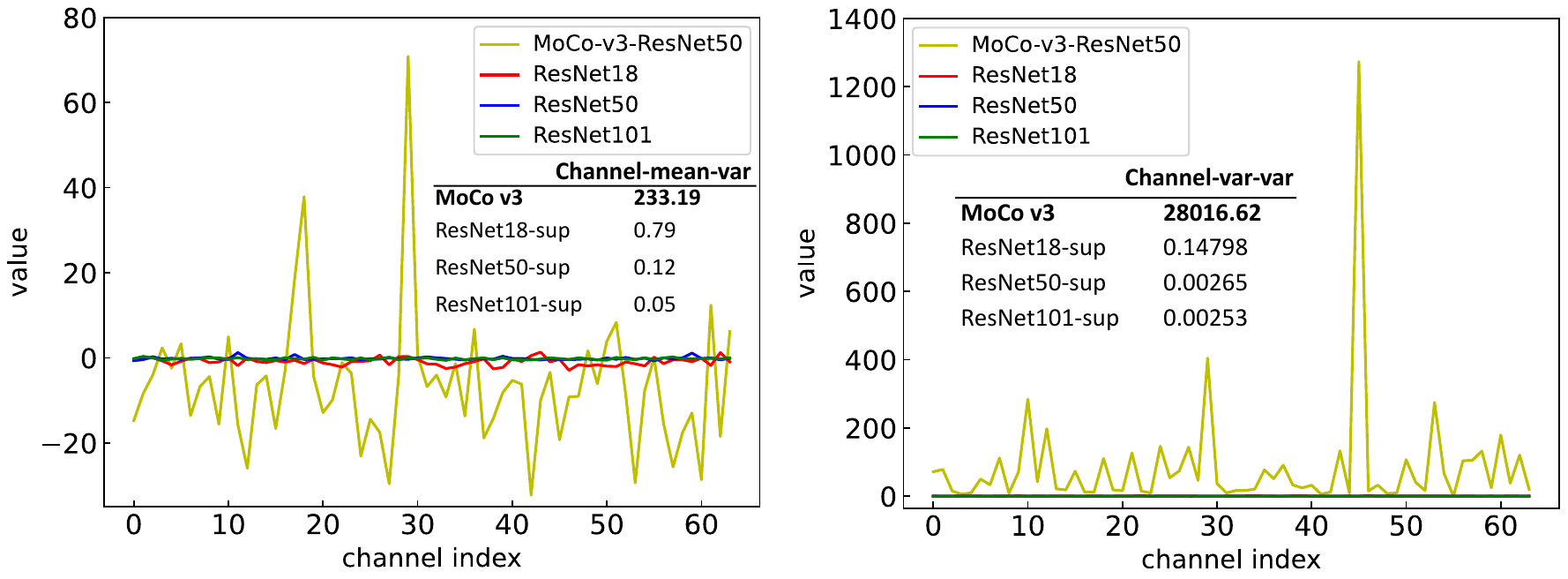}
		\vspace{-0.15in}
		\caption{Illustration of mean (left) and variance (right) of the first BN layer in the residual block from self-supervised MoCo-v3-ResNet-50, supervised ResNet-\{18, 50, 101\}. In each subfigure, the x-axis represents the channel index, y-axis represents the corresponding value. The table inside each subfigure represents the variance across all channels, which reflects the fluctuation of statistics in the BN layer.}
		\label{fig:self-supervised}
		\vspace{-0.15in}
	\end{figure*}
	
	\vspace{-0.08in}
	\begin{theorem} Batch Normalization statistical parameters $\Theta$ (mean $\mu$ and variance $\sigma^2$) derived from self-supervised contrastive learning are more fluctuant than those from supervised learning, which is more informative for dataset distillation recovery of image synthesis with higher entropy, i.e., \( H(\Theta_{ssl}) \!>\! H(\Theta_{sl}) \).
	\end{theorem}
	\vspace{-0.05in}
	
	\noindent{\bf Problem statement}: Let \( \Theta_{ssl} \) denote the batch norm parameters (mean \( \mu_{ssl} \) and variance \( \sigma_{ssl}^2 \)) obtained from a model trained using self-supervised contrastive learning, and let \( \Theta_{sl} \) denote the batch norm parameters (mean \( \mu_{sl} \) and variance \( \sigma_{sl}^2 \)) obtained from a model trained using traditional supervised learning. We claim that \( \Theta_{ssl} \) encapsulates a richer representation of the underlying data distribution compared to \( \Theta_{sl} \). 
	For a given dataset $\mathcal D_l$ with data samples $x_i\!\in\!\mathcal D_l$, the batch norm parameters for self-supervised contrastive learning and supervised learning are as follows: 
	1) For self-supervised contrastive learning:
	\vspace{-0.08in}
	\begin{equation}
		\mu_{ssl} = \frac{1}{|\mathcal D_l|} \sum_{i=1}^{|\mathcal D_l|} f_{\theta}(x_i), \sigma_{ssl}^2 = \frac{1}{|\mathcal D_l|} \sum_{i=1}^{|\mathcal D_l|} (f_{\theta}(x_i) - \mu_{ssl})^2 
	\end{equation}
	
	2) For supervised learning:
	\vspace{-0.05in}
	\begin{equation}
		\mu_{sl} = \frac{1}{|\mathcal D_l|} \sum_{i=1}^{|\mathcal D_l|} f_{\phi}(x_i), \sigma_{sl}^2 = \frac{1}{|\mathcal D_l|} \sum_{i=1}^{|\mathcal D_l|} (f_{\phi}(x_i) - \mu_{sl})^2 
		\vspace{-0.05in}
	\end{equation}
	where \( f_{\theta} \) and \( f_{\phi} \) are the feature-extracting functions of the models trained with self-supervised contrastive learning and supervised learning, respectively.
	
	\noindent{\bf \em Proof}: To prove that \( \Theta_{ssl} \) is more informative than \( \Theta_{sl} \), we can analyze the entropy of the resulting feature distributions. The entropy of a distribution is a measure of its information content, with higher entropy indicating a more informative distribution. 
	The entropy \( H \) of the batch-normalized features can be defined as:
	\vspace{-0.05in}
	\begin{equation}
		H(\Theta) = - \int p_\Theta(x) \log p_\Theta(x) dx 
		\vspace{-0.05in}
	\end{equation}
	where \( p_\Theta(x) \) is the probability density function of the batch-normalized features.
	
	Assuming that the amount of pretraining data is large and diverse enough, according to the central limit theorem~\cite{rosenblatt1956central}, the features follow a Gaussian distribution post-normalization, we can simplify the entropy of both distributions to (detailed proof is provided in Appendix.):
	\vspace{-0.08in}
	\begin{equation}
		H(\Theta_{ssl}) = \frac{1}{2} \log(2 \pi e \sigma_{ssl}^2), H(\Theta_{sl}) = \frac{1}{2} \log(2 \pi e \sigma_{sl}^2)
	\end{equation}
	This expression gives us the entropy of a Gaussian distribution in terms of its variance $\sigma^2$.  
	The entropy is maximized when the variance is large, indicating that a broader distribution (more uncertainty) leads to higher entropy. 
	
	Given that self-supervised contrastive objective encourages a model to learn an embedding space where similar samples are close to each other and dissimilar ones are further apart, this process preserves more intrinsic data variability compared to supervised learning which may collapse representations to discriminative features relevant only to class labels.
	
	Consequently, we hypothesize that \( \sigma_{ssl}^2 \!>\! \sigma_{sl}^2 \), leading to \( H(\Theta_{ssl}) \!>\! H(\Theta_{sl}) \). To empirically validate this hypothesis, one would conduct extensive experiments to compare the variances of features obtained through both learning methods across various datasets, as shown in Fig.~\ref{fig:self-supervised}. 
	
	Hence, under the assumption that self-supervised contrastive learning leads to a greater preservation of data variability, the batch normalization parameters derived from this approach can be considered more informative than those derived from supervised learning, as evidenced by the greater entropy in the batch-normalized feature distribution.
	
	\noindent{\bf How to Choose Optimal Pretrained Models.} One of our key contributions is the proposed selection criterion for the BN-matching-based dataset distillation frameworks. Through theoretical proof and empirical experimental phenomenon analysis, we clarify that the degree of variation in these BN statistics is the key for enjoying the large capability of the BN-matching data synthesis.
	
	\begin{figure}[h]
		\centering
		\vspace{-0.25in}
		\includegraphics[width=0.48\linewidth]{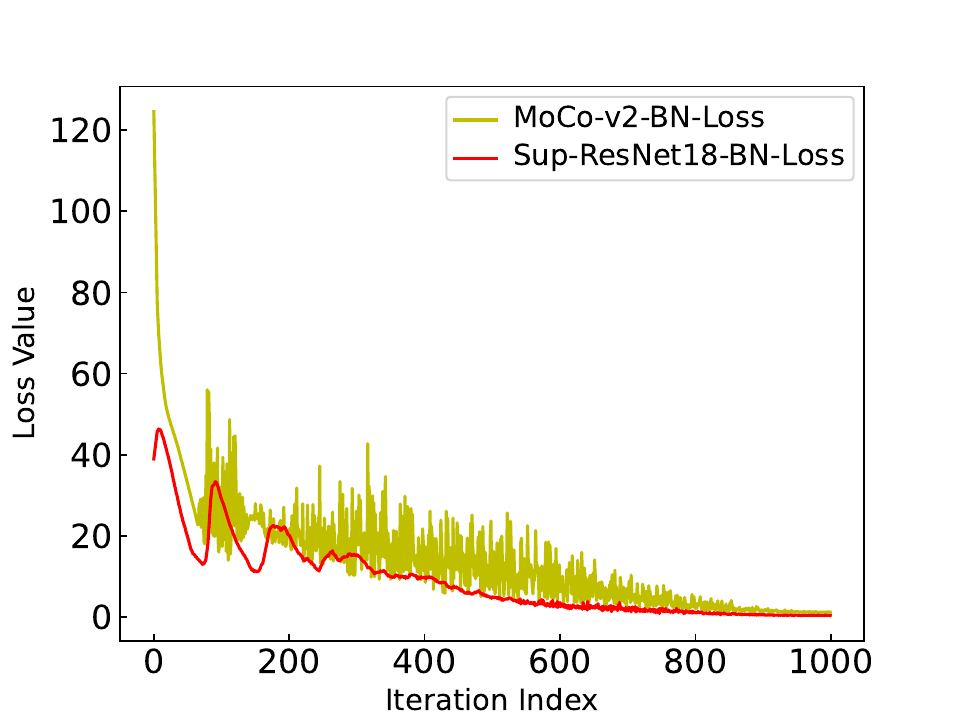}
		\includegraphics[width=0.46\linewidth]{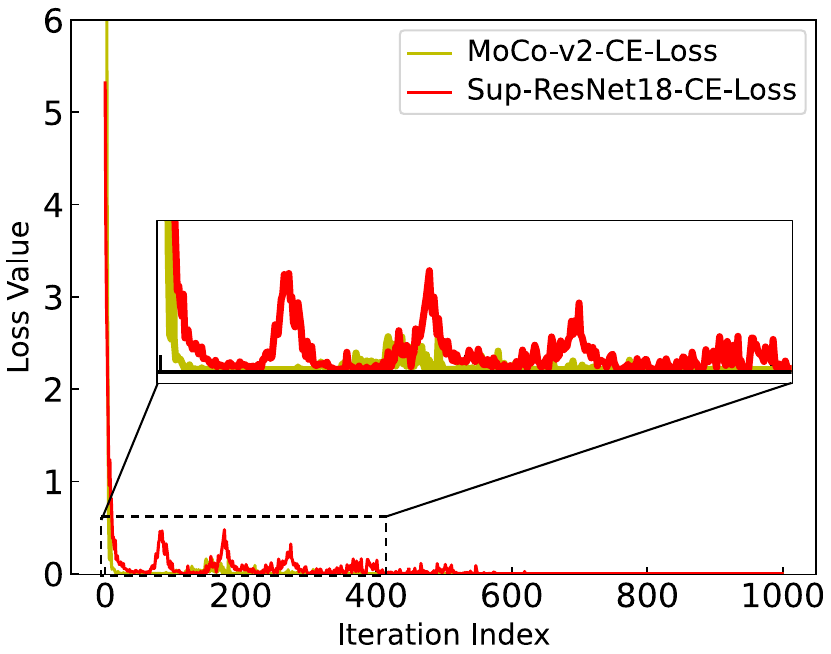}
		\vspace{-0.1in}
		\caption{Loss trajectories during data synthesis. Left subfigure illustrates the BN loss term and right subfigure illustrates the CE loss term. The backbone is ResNet-18 for both self-supervised and supervised training schemes.} 
		\label{fig:trajectories}
		\vspace{-0.2in}
	\end{figure}
	
	\noindent{\bf Optimization Effects.} We explore how the pertaining scheme and BN statistic distributions affect the optimization during data recovery.  As shown in Eq.~\ref{recovery} of the following Sec.~\ref{recovery_loss}, our objective for recovery is $\mathcal{L}_\text{CE}+\mathcal{L}_\text{BN}$ matching. To understand the optimization effects using this loss combination, we visualize the individual loss trajectories for both $\mathcal{L}_\text{CE}$ and $\mathcal{L}_\text{BN}$. As shown in Fig.~\ref{fig:trajectories}, we have two interesting observations: Firstly, from left subfigure we can see that when employing the self-supervised pretrained model, larger $\mathcal{L}_\text{BN}$ loss at the early stage occurs which is naturally expected as the variance of BN statistics is larger upon the discussion above, while the final loss magnitude is similar to the ones that are from the supervised model, this phenomenon aligns with our conjecture that the variance of BN supervision helps optimize the synthetic data in early and middle stages more thoroughly, leading to higher generation quality. More interestingly, the second observation from the right subfigure shows that the varied BN supervisions can even stabilize the loss curve of the main $\mathcal{L}_\text{CE}$ term. We emphasize that these two findings have not been found or discussed by any prior literature in the dataset distillation task. 
	
	\vspace{-0.1in}
	\subsection{Model-Data Alignment}
	\vspace{-0.04in}
	
	Finetuning~\cite{yosinski2014transferable,shen2021partial} is a common practice for adapting a pretrained model to specific target data. However, this method alters the pretrained model's parameters $\Theta_{ssl}$, potentially disrupting the original BN distributions. Linear probing is an alternative way for aligning pretrained model to the target dataset. We found it is more effective than fine-tuning for data synthesis, which has also demonstrated superiority in other tasks like domain adaptation \cite{kumar2022fine,kirichenko2022last}. Given that $f_{v, w}(x)\!=\!v^{\top} f(x)$, our approach involves keeping $f$ frozen while focusing on learning $v$. This strategy effectively separates the learning of intermediate feature distributions from the alignment of higher-level semantic information.
	
	\vspace{-0.08in}
	\subsection{Imbalanced BN Statistic Distribution Matching}
	\vspace{-0.04in}
	
	In the data synthesis phase, we use current batch's mean and variance statistics to match the self-supervised pretrained global statistics. Unlike SRe$^2$L, we apply new imbalanced coefficients according to the property of pretrained models as the objective, which can be formulated as an imbalanced {\em batch-match-global} manner:
	\begin{equation} \label{BN_coefficient}
		\small
		\begin{aligned}
			\mathcal{R}_{\mathrm{reg}}(\boldsymbol{x}^{\prime}) & \!=\!\sum_k\!\beta_k\left\|\mu_k(\boldsymbol{x}^{\prime})\!-\!\mathbb{E}\!\left(\mu_k \!\mid\! \mathcal{D}_k\right)\right\|_2\!+\!\sum_k\!\gamma_k\left\|\sigma_k^2(\boldsymbol{x}^{\prime})\!-\!\mathbb{E}\!\left(\sigma_k^2 \!\mid\! \mathcal{D}_k\right)\right\|_2 \\
			& \!\approx \! \sum_k\beta_k\left\|\mu_k(\boldsymbol{x}^{\prime})\!-\!\mathbf{B N}_k^{\mathrm{RM}}\right\|_2 \!+\!\sum_k\gamma_k\left\|\sigma_k^2(\boldsymbol{x}^{\prime})\!-\!\mathbf{B N}_k^{\mathrm{RV}}\right\|_2
		\end{aligned} 
	\end{equation}
	where $k$ is the index of $\mathrm{BN}$ layer, $\mu_k(\boldsymbol{x}^{\prime})$ and $\sigma_k^2(\boldsymbol{x}^{\prime})$ are the channel-wise mean and variance in current batch data. $\mathbf{BN}_k^{\mathrm{RM}}$ and $\mathbf{B N}_k^{\mathrm{RV}}$ are mean and variance in the pretrained model at $k$-th $\mathrm{BN}$ layer, which are globally counted. $\beta_k$ and $\gamma_k$ are coefficients to control the contributions of different layers. In our experiments, we also tried to make these hyperparameters to be learnable and the accuracy is similar to our grid search experiments in Table~\ref{tab:BN_para}.
	
	\vspace{-0.1in}
	\subsection{A Simple DD Framework via Self-supervised Pretraining} \label{recovery_loss}
	\vspace{-0.05in}
	
	As illustrated in Fig.~\ref{overview}, here we summarize how to develop a strong data distillation baseline by integrating insights and methods from above analyses alongside established optimization strategies. We show that our simple framework already achieves state-of-the-art accuracy using the same evaluation protocol. This can be a crucial contribution towards understanding the true impact of the method for dataset distillation and towards minimizing the true gap between the distilled datasets and full datasets. 
	We leverage self-supervised learned models and apply the data synthesis objective as follows:
	\vspace{-0.06in}
	\begin{equation} \label{recovery}
		\underset{\mathcal{D}_{{d}},|\mathcal{D}_d|}{\arg \min } \ \ell\left(\mathcal{M}_{\boldsymbol{\theta}_{\mathcal{D}_l}}\left({\boldsymbol{x}}^{\prime}_{\mathrm{syn}}\right), \boldsymbol{y}\right)+\alpha\mathcal{R}_{\mathrm{reg}}
		\vspace{-0.05in}
	\end{equation}
	where $\mathcal{M}_{\boldsymbol{\theta}_{\mathcal{D}_l}}$ is the SSL pretrained model after alignment. The first term is the cross-entropy loss with ground-truth, and the second is the BN matching loss. $\alpha$ is the coefficient to balance the contributions of these two losses.
	
	\vspace{-0.1in}
	\subsection{Post-training for Validation} 
	\vspace{-0.05in}
	
	Similar to~\cite{yin2023squeeze}, a pre-generated soft label~\cite{shen2022fast} scheme is employed to eliminate the teacher models trained on the original dataset in post-validation. The post-training objective is:
	\vspace{-0.09in}
	\begin{equation}
		\mathcal{L}_{\mathrm{syn}}=-\sum_i^{|\mathcal D_d|} {\boldsymbol{y}}_i^{\prime} \log \phi_{\boldsymbol{\theta}_{\mathcal{D}_{{d}}}}\left({\boldsymbol{x}}^{\prime}_{\mathbf{R}_i}\right)
		\vspace{-0.06in}
	\end{equation}
	where ${\boldsymbol{x}}^{\prime}_{\mathbf{R}_i}$ is the $i$-th crop in the synthetic image and ${\boldsymbol{y}}_i^{\prime}$ is the corresponding soft label. ${\boldsymbol{y}}_i^{\prime}$ is from the SSL pretrained model after alignment. Finally, we can train the model $\phi_{\boldsymbol{\theta}_{\mathcal{D}_{{d}}}}$ on the synthetic data using the objective.
	
	\vspace{-0.08in}
	\section{Experiments}
	
	\vspace{-0.06in}
	\subsection{Datasets and Implementation Details}
	We verify the effectiveness of our approach on various datasets, including CIFAR-100~\cite{krizhevsky2009learning}, Tiny-ImageNet~\cite{le2015tiny}, and ImageNet-1K~\cite{deng2009imagenet}. \textbf{CIFAR-100} contains 50K images with 32$\times$32 pixels categorized into 100 classes. \textbf{Tiny ImageNet} comprises 200 classes with 500 images of 64$\times$64 resolution per class. 
	\textbf{ImageNet-1K} consists of approximately 1.2M training images with 224$\times$224 resolution into 1000 classes. We squeeze training datasets based on the self-supervised pretraining and evaluate the performance of synthetic data using original validation datasets. Except the self-supervised pertaining, recovery and post-training configurations follow the protocol of SRe$^2$L~\cite{yin2023squeeze}. Specifically, we use fewer 3K iterations in recovery according to our ablations in Table~\ref{tab:moco_v3_3K} instead of 4K as SRe$^2$L used. The post-training budgets on various datasets are the same as SRe$^2$L~\cite{yin2023squeeze} and we examine more network structures for the cross-architecture evaluation, such as RegNet-X-8gf~\cite{radosavovic2020designing}, SqueezeNetV1.0~\cite{iandola2016squeezenet}, MobileNet-V3-L~\cite{howard2019searching}, and Shuffl-eNetV2-0.5x~\cite{ma2018shufflenet}. The computational resources employed in these experiments include the NVIDIA A100 (40G) and 4090 GPUs. More details (e.g., $\alpha$) regarding our experimental settings can be referred to our Appendix. 
	
	\vspace{-0.08in}
	\subsection{Comparison with State-of-the-art Approaches}
	\vspace{-0.04in}
	In this section, we compare the performance of our approach with current state-of-the-art methods in Table \ref{tab:imagenet}. 
	These results demonstrate the effectiveness of our method in various large-scale dataset distillation scenario.
	
	\noindent{\bf CIFAR-100.} The results are presented in the first group of Table \ref{tab:imagenet}. MoCo-v2-200ep ResNet-18 is employed as the compression model, with 
	the same architecture ResNet-18 used in SRe$^2$L serving as the validation model. 
	Our approach achieves a significant 4.0\% enhancement over SRe$^2$L when the IPC is set to 50.
	
	\noindent{\bf Tiny-ImageNet.} The results of Tiny-ImageNet in the second group of Table \ref{tab:imagenet} show that our approach consistently outperforms SRe$^2$L, achieving a higher validation accuracy with an improvement of 4.8\% under the IPC setting of 50. 
	
	\begin{table*}[t]
\renewcommand\arraystretch{0.9}
\centering
\scriptsize
\setlength{\tabcolsep}{2pt}
\caption{Comparison of dataset distillation and coreset selection methods. We trained 3 times each to get $\bar{m}\pm n$. * indicates the result is not provided. SRe$^2$L~\cite{yin2023squeeze}, ours and whole data results are based on the ResNet-18 as the backbone.}
\label{tab:imagenet}
\vspace{-9pt}
\resizebox{0.999\linewidth}{!}{
\begin{tabular}{ccc|rrr|rccccl|c}
\toprule
\multirow{3}{*}{}           & \multirow{2}{*}{IPC} & \multirow{2}{*}{Ratio \%} & \multicolumn{3}{c|}{Coreset Selection}    & \multicolumn{6}{c|} {Training Set Synthesis} & Whole \\ %
                            & &                & \multicolumn{1}{c}{Random}        & \multicolumn{1}{c}{Herding}       & \multicolumn{1}{c|}{Forgetting}     			& \multicolumn{1}{c}{\texttt{DM} \cite{zhao2022dataset}}                &\multicolumn{1}{c}{\texttt{CAFE+DSA} \cite{wang2022cafe}}         & \multicolumn{1}{c}{\texttt{MTT}~\cite{cazenavette2022dataset}}  & \multicolumn{1}{c}{\texttt{TESLA}~\cite{cui2023scaling}} & \multicolumn{1}{c}{\texttt{SRe$^2$L}~\cite{yin2023squeeze}} &  \multicolumn{1}{l|}{ \ \ \bf{Ours}}  & Data  \\ \midrule
                   
\multirow{3}{*}{CIFAR-100}     & 1   & 0.2    &  4.2 $\pm$ 0.3  &  8.4 $\pm$ 0.3  &  4.5 $\pm$ 0.2      	& 11.4 $\pm$ 0.3  & 14.0 $\pm$ 0.3 &  {24.3 $\pm$ 0.3} &24.8$\pm$0.4 &\multicolumn{1}{c}{-}  & \ \ \ \ \ \ \ - & \multirow{3}{*}{79.1}  \\ 
                              & 10  & 2      & 14.6 $\pm$ 0.5  & 17.3 $\pm$ 0.3  & 15.1 $\pm$ 0.3   	& 29.7 $\pm$ 0.3   & 31.5 $\pm$ 0.2 &   {40.1 $\pm$ 0.4\;}   &41.7$\pm$0.3   &\multicolumn{1}{c}{-}  & \ \ \ \ \ \ \ -   \\  
                              & 50  & 10     & 30.0 $\pm$ 0.4  & 33.7 $\pm$ 0.5  & 30.5 $\pm$ 0.3                                   & 43.6 $\pm$ 0.4   & 42.9 $\pm$ 0.2 &      {47.7 $\pm$  0.2}    &47.9$\pm$0.3 &49.4$\pm$*  &  \bf 53.4$\pm$1.1  \\  \midrule

\multirow{3}{*}{Tiny ImageNet} & 1   & 0.2    &  1.4 $\pm$ 0.1  &  2.8 $\pm$ 0.2  &  1.6 $\pm$ 0.1              		            &  3.9 $\pm$ 0.2  & \multicolumn{1}{c}{-}  &{8.8 $\pm$ 0.3\;}  &\multicolumn{1}{c}{-} & \multicolumn{1}{c}{-} &  \ \ \ \ \ \ \ \  -  & \multirow{3}{*}{61.2}  \\ 
                              & 10  & 2      &  5.0 $\pm$ 0.2  &  6.3 $\pm$ 0.2  &  5.1 $\pm$ 0.2               & 12.9 $\pm$ 0.4   & \multicolumn{1}{c}{-} &{23.2 $\pm$ 0.2\;}   &\multicolumn{1}{c}{-} &\multicolumn{1}{c}{-} & \bf 31.6$\pm$0.1   \\  
                              & 50  & 10     & 15.0 $\pm$ 0.4  & 16.7 $\pm$ 0.3  & 15.0 $\pm$ 0.3         &24.1 $\pm$ 0.3  & \multicolumn{1}{c}{-} &{28.0 $\pm$ 0.3\;}  &\multicolumn{1}{c}{-}  & 41.1$\pm$0.4  & \bf 45.9$\pm$0.2 \\ \midrule
 \multirow{4}{*}{ImageNet-1K}    &10  & 0.8   &\multicolumn{1}{c}{-}   &\multicolumn{1}{c}{-}   & \multicolumn{1}{c}{-} & \multicolumn{1}{c}{-} & \multicolumn{1}{c}{-} & \multicolumn{1}{c}{-} &17.8$\pm$1.3 & 21.3$\pm$0.6  & \bf{32.1$\pm$0.2}  & \multirow{4}{*}{73.2} \\
                            &50  & 3.9   & \multicolumn{1}{c}{-}    & \multicolumn{1}{c}{-} & \multicolumn{1}{c}{-} & \multicolumn{1}{c}{-} & \multicolumn{1}{c}{-} & \multicolumn{1}{c}{-} &27.9$\pm$1.2 & 46.8$\pm$0.2 &  \textbf{53.1$\pm$0.1} \\
                             &100  & 7.8  &\multicolumn{1}{c}{-}    & \multicolumn{1}{c}{-} & \multicolumn{1}{c}{-} & \multicolumn{1}{c}{-} & \multicolumn{1}{c}{-} & \multicolumn{1}{c}{-} & \multicolumn{1}{c}{} &\multicolumn{1}{c}{52.8$\pm$0.3} &  \textbf{57.9$\pm$0.1}  \\  
                             &200  & 11.7  &\multicolumn{1}{c}{-}    & \multicolumn{1}{c}{-} & \multicolumn{1}{c}{-} & \multicolumn{1}{c}{-} & \multicolumn{1}{c}{-} & \multicolumn{1}{c}{-} & \multicolumn{1}{c}{} &\multicolumn{1}{c}{57.0$\pm$0.4} &  \textbf{63.5$\pm$0.1}  \\ 
    
\bottomrule

\end{tabular}
}
\vspace{-0.2in}
\end{table*}

	\noindent{\bf ImageNet-1K.} As shown in the third group of Table \ref{tab:imagenet}, our method demonstrates notable improvements compared to SRe$^2$L, with increases of 10.8\%, 6.3\%, 5.1\%, and 6.5\% under IPC 10, 50, 100, and 200, respectively. A more comprehensive comparison on various post-training architectures and IPCs is shown in Table~\ref{tab:new_comp}, which demonstrates significant improvements over prior art SRe$^2$L.
	
	\vspace{-0.12in}
	\subsection{Ablation}
	\vspace{-0.06in}
	\label{sec:Ablation}
	
	Our framework encompasses three core phases: compression, recovery, and validation. We carry out extensive ablation experiments in each phase to assess the various factors influencing the overall results.
	
	\begin{wrapfigure}{l}{0.52\textwidth}
		\centering
		\vspace{-0.25in}
		\includegraphics[width=0.85\linewidth]{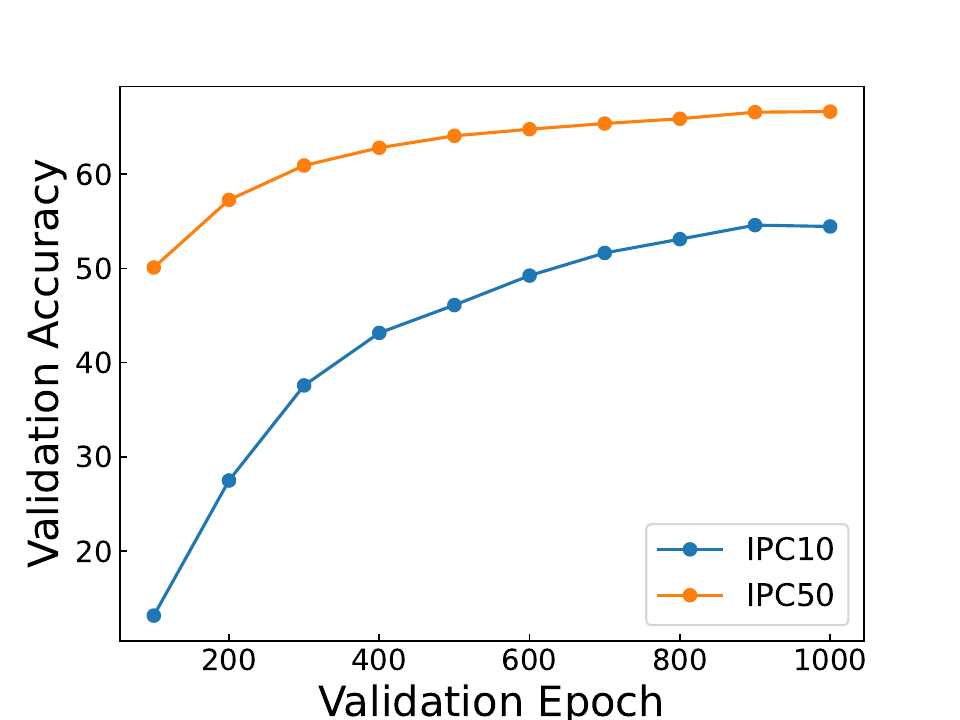}
		\vspace{-0.13in}
		\caption{Top-1 accuracy on ImageNet-1K distilled images from MoCo v3 ResNet-50 under IPC 10 and 50 across various validation epochs. Post-validation models of ResNet-18 are saturated after 800 epochs in validation.}
		\label{fig:training_budget}
		\vspace{-0.25in}
	\end{wrapfigure}
	
	\noindent{\bf{Compression Phase}}. 
	(1) {\bf Self-supervised Pretraining Budget:} We examine the impact of the number of pretraining epochs in the compression phase, showcasing the validation accuracy across three models: ResNet-\{18, 50, 101\}. 
	As shown in each group of Table \ref{tab:moco_v2_ep}, longer pretraining does not necessarily lead to better results. For instance, the optimal performance for MoCo v2 is observed at 200 epochs, aligning with our earlier assessments and discussions for compression that the variation in BN statistics is the key for the BN-matching-based data synthesis, instead of the higher pretraining performance. We also show the channel-wise variances of different self-supervised models' first layer BN mean in the last column, which aligns with the final performance well. 
	{(2) \textbf{Architecture}}: We examine the effects of various pretrained model sizes and differing model widths on compression. (i) {\bf Model Size}: Our analysis, as shown in the first group of Table \ref{tab:moco_v2_model}, reveals that larger pretrained models generally can lead to higher validation accuracy. This finding supports the strategy of scaling up the pretrained model size, a trend that is increasingly popular in current practices. Such scaling proves beneficial, significantly enhancing our model's performance. (ii) {\bf Model Width}: The second group of Table \ref{tab:moco_v2_model} indicates that while expanding the width of pretrained ResNet-50 model can lead to improved accuracy, widening it to w2 appears to be sufficient. This width demonstrates a performance nearly equivalent to that of w4 and w5 models.
	
	\begin{table*}[t]
\vspace{-0.2in}
\begin{minipage}{0.5\textwidth}
\centering\small
\setlength{\tabcolsep}{2pt}
\caption{Validation results on distilled ImageNet-1K datasets recovered from different pretrained models and epochs. Recovery models are ResNet-50 MoCo v2~\cite{chen2020improved}, MoCo v3~\cite{chen2021empirical}, SwAV~\cite{caron2021unsupervised}, and DINO~\cite{caron2021emerging}. }
\label{tab:moco_v2_ep}
\vspace{-0.1in}
\resizebox{0.99\linewidth}{!}{
\begin{tabular}{@{}ccccc@{}|c}
\toprule
& \multicolumn{1}{c}{\multirow{1}{*}{pretrain}} & \multicolumn{4}{c}{validation accuracy (\%)} \\ \cline{3-6} 
\multicolumn{1}{c}{}   & \multicolumn{1}{c}{\multirow{1}{*}{epoch}}                & Res-18 & Res-50 & Res-101 & BN-mean Var \\ \midrule
\multirow{3}{*}{{MoCo v2~\cite{chen2020improved}}}    & 200 & \bf 46.62   & \bf 56.72   & \bf 56.60 & \bf 5.98$\times$10$^-3$ \\ 
                                         & 400 &  45.51   &  55.12   &  55.93  & 4.94$\times$10$^-3$\\  
                                         & 800 &   39.35  & 49.93    &  50.28 & 3.00$\times$10$^-3$ \\ 
 \midrule
\multirow{3}{*}{{MoCo v3~\cite{chen2021empirical}}} &100   & 49.75       & 57.95       & 57.24  & 2.16$\times$10$^2$   \\
 & 300   & \bf 53.13       & \bf 60.92       & \bf 60.98  & \bf 2.33$\times$10$^2$  \\
 & 1000 & 50.62       & 59.21       & 59.23  & -   \\ 
\midrule
\multirow{4}{*}{{SwAV~\cite{caron2021unsupervised}}} & 100 &  34.72   &   44.96  &   44.24  & 2.03$\times$10$^2$ \\
 & 200 &   47.55  &   57.32  &  56.26 & 2.91$\times$10$^2$   \\
 & 400 &   48.02  &   58.25  &  \bf 57.90  & \bf 3.82$\times$10$^2$  \\
 & 800 & \bf 49.70    & \bf 58.70    & 56.39   & 3.40$\times$10$^2$   \\ 
\midrule
\multirow{2}{*}{{DINO~\cite{caron2021emerging}}} & 100 &  46.08   & \bf 55.32   & \bf 55.41 & -  \\
 & 800 & \bf 47.27    & 51.73    & 54.40   & -  \\ 
 \bottomrule
\end{tabular}
}
\end{minipage}
 \hfill
\begin{minipage}{0.48\textwidth}
\vspace{0.35in}
\centering
\caption{ImageNet-1K val results on distilled datasets from MoCo v2 and SwAV.}
\label{tab:moco_v2_model}
\vspace{-0.1in}
\resizebox{0.97\linewidth}{!}{
\begin{tabular}{@{}ccccc@{}}
\toprule
\multicolumn{2}{c}{\multirow{2}{*}{pretrain model}} & \multicolumn{3}{c}{validation accuracy (\%)} \\ \cline{3-5} 
\multicolumn{2}{c}{}                   & ResNet-18 & ResNet-50 & ResNet-101 \\ 
\midrule
\multirow{3}{*}{\rotatebox{90}{MoCov2}}  & ResNet-18  &  43.15 & 53.41 &  53.72 \\
 & ResNet-50  &  46.62   &  56.72   &  56.60   \\
 & ResNet-101 &   \bf  47.71 &   \bf  57.12 &   \bf   56.88 \\ 
\midrule
\multirow{4}{*}{\rotatebox{90}{SwAV}}  & ResNet-50    & 48.02  &   58.25  &  57.90 \\
 & ResNet-50-w2 &  49.80   & \bf 58.93   &  59.51  \\
 & ResNet-50-w4 & \bf 49.83   &  58.72   &  58.46     \\
 & ResNet-50-w5 &  49.59   &  58.80   & \bf 59.70    \\ \bottomrule
\end{tabular}
}
\vspace{0.08in}
\caption{Validation results on distilled ImageNet-1K datasets recovered from MoCo v3 with various recovery iterations.}
\label{tab:moco_v3_3K}
\vspace{-0.1in}
\resizebox{0.98\linewidth}{!}{
\begin{tabular}{@{}cccc@{}}
\toprule
\multirow{2}{*}{\begin{tabular}[c]{@{}c@{}}MoCo v3 \\recovery iteration\end{tabular}} & \multicolumn{3}{c}{validation accuracy (\%)} \\ \cmidrule(l){2-4} 
& ResNet-18 & ResNet-50  & ResNet-101 \\ \midrule
1K &  51.36  & 59.78   & 60.06   \\
2K &  52.97  & 60.63   & 60.09   \\
3K & \bf 53.13  & \bf 60.92   & \bf 60.98     \\ 
4K &  51.90  & 59.98   & 60.35  \\\bottomrule
\end{tabular}
} 
\vspace{-0.15in}

\end{minipage}

\end{table*}

	\begin{table*}[t]
\vspace{-0.2in}
\begin{minipage}{0.52\textwidth}
\vspace{-0.2in}
\centering
\caption{Ablation on the coefficient of the first BN loss term in Eq.~\ref{BN_coefficient} when $k=0$, other layers' coefficients are set to 1 following~\cite{yin2023squeeze}. The validation results are derived from the distilled ImageNet-1K data with IPC 50 from MoCo-v3-300ep ResNet-50 in recovery, and ResNet-\{18, 50, 101\} in post-validation.} 
\label{tab:BN_para}
\vspace{-0.1in}
\resizebox{0.99\linewidth}{!}{
\begin{tabular}{@{}cccc@{}}
\toprule
\multirow{2}{*}{\begin{tabular}[c]{@{}c@{}}Frist BN multiplier\\ recovery parameter\end{tabular}} & \multicolumn{3}{c}{validation accuracy (\%)} \\ \cmidrule(l){2-4} 
         & ResNet-18 & ResNet-50 & ResNet-101 \\ \midrule
1$\times$       &  52.49   &  60.10    & 58.98   \\
5$\times$       &  53.09   &  60.43    & \bf 61.11  \\
10$\times$      &  \bf 53.13   & \bf 60.92    &  60.98    \\
15$\times$      &  50.96   &  58.15    & 56.29  \\
20$\times$      &  50.82   &  57.69   &  57.30  \\ \bottomrule
\end{tabular}
}
\end{minipage}
 \hfill
\begin{minipage}{0.47\textwidth}
\vspace{0.28in}
\caption{Validation result comparison with SRe$^2$L~\cite{yin2023squeeze} on four different IPCs and three model architectures.}
\label{tab:new_comp}
\vspace{-0.1in}
\resizebox{0.99\linewidth}{!}{
\begin{tabular}{llcccc}
\toprule
\multirow{2}{*}{Network}    & \multirow{2}{*}{Method} & \multicolumn{4}{l}{validation accuracy (\%)} \\ \cline{3-6} 
                            &                         & IPC=10       & 50      & 100      & 200      \\ \midrule
\multirow{2}{*}{ResNet-18}  & SRe$^2$L~\cite{yin2023squeeze}       &   21.3 &  46.8 &  52.8 & 57.0  \\
                            & Ours                                 &   \bf 32.1   & \bf 53.1  & \bf 57.9  & \bf 63.5    \\ \midrule
\multirow{2}{*}{ResNet-50}  & SRe$^2$L~\cite{yin2023squeeze}       &  28.4   &  55.6  &  61.0 & 64.6 \\
                            & Ours                                 & \bf 38.9    & \bf 60.9   &  \bf 65.8  & \bf 67.8   \\ \midrule
\multirow{2}{*}{ResNet-101} & SRe$^2$L~\cite{yin2023squeeze}       &   30.9   &  60.8 &  62.8 & 65.9   \\
                            & Ours                                 &  \bf 39.6    & \bf 61.0  &  \bf 65.6   &  \bf 68.2  \\ \bottomrule
\end{tabular}
}
\end{minipage}

\vspace{-0.1in}
\end{table*}
	
	\noindent\textbf{Recovery Phase}. 
	(1) \textbf{BN coeffcient}: The term BN coeffcient refers to the {\em first batch-norm multiplier} parameter. As shown in Table \ref{tab:BN_para}, two among three of our highest accuracy are achieved with a BN setting of 10. Consequently, we adopt 10 for all subsequent experiments. It is important to note that the marked change in accuracy as BN values are adjusted. This transition, from rising to declining accuracy with increasing BN values, underscores the importance of carefully fine-tuning the BN coefficient to optimize model performance. 
	(2) {\bf Iteration Budget}: 
	As in Table \ref{tab:moco_v3_3K}, we observe the optimal performance at 3K iterations with a noticeable turning point in accuracy occurring at this mark. This recovery budget is smaller than SRe$^2$L while achieving higher accuracy, indicating that simply increasing iterations does not linearly correlate with enhanced performance without employing a better synthesis approach. It also emphasizes the importance of a strategic approach in calibrating the iteration count. 
	
	\noindent\textbf{Post-evaluation Phase}. In this phase, we assess the effectiveness of various validation models, specifically ResNet-18, 50, and 101, in evaluating the quality of distilled images. As presented in different groups of Table \ref{tab:moco_v2_ep}, a general trend is observed where accuracy improves as the size of the model increases. However, it is particularly notable that ResNet-50 and ResNet-101 demonstrate similar performance levels. This trend is observed to be consistently maintained across various experimental setups, despite any changes in the experimental conditions. Moreover, as shown in Fig.~\ref{fig:training_budget}, during the post-validation stage, we observe that longer training on the distilled images leads to higher accuracy, with significant enhancements across various IPC settings. However, it is important to note that models tend to reach a saturation point after approximately 800 epochs.
	
	\vspace{-0.08in}
	\subsection{Analysis}
	\vspace{-0.04in}

\begin{figure*}[t]
		\begin{minipage}{0.56\textwidth}
			\centering
			\includegraphics[width=0.99\linewidth]{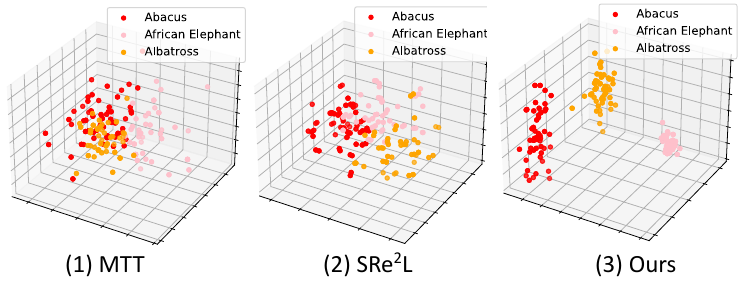}
			\vspace{-0.25in}
			\caption{Synthetic data clustering of MTT, SRe$^2$L and ours on Tiny-ImageNet with three classes: Abacus, African Elephant and Albatross.}
			\label{fig:clustering}
			\vspace{-0.14in}
		\end{minipage}
		\hfill
		\begin{minipage}{0.43\textwidth}
			\centering
			\vspace{-0.14in}
			\captionof{table}{Validation results using different datasets with distilled data from MoCo v2 pretrained models.}
			\label{tab:cross_dataset}
			\resizebox{1.01\linewidth}{!}{
				\begin{tabular}{@{}cccc@{}}
					\toprule
					\multirow{2}{*}{\begin{tabular}[c]{@{}c@{}}MoCo v2\\ pretraining \end{tabular}} & \multicolumn{3}{c}{validation accuracy (\%)} \\ \cmidrule(l){2-4} 
					& ResNet-18 & ResNet-50 & ResNet-101 \\ \midrule
					ImageNet-1K   &  46.62   &  56.72   &  56.60   \\
					ImageNet-21K  &  40.55   & 50.34    &  48.71  \\ \midrule
					CIFAR-100     &  56.63   &  58.67   &   58.00     \\
					Tiny-ImageNet &  46.43   &  47.85   &  49.14  \\
					\bottomrule
				\end{tabular}
			} 
			\vspace{-0.15in}
		\end{minipage}

	\end{figure*}

	\noindent{\bf Synthetic Image Clustering.} We perform {\em kmeans}~\cite{macqueen1967some} on the synthetic images with PCA~\cite{jolliffe2016principal} (to reduce the input pixels to three dimensions) for obtaining the clustering distributions, as illustrated in Fig.~\ref{fig:clustering}. It reflects that our synthetic images have the best semantic capability than MTT and SRe$^2$L.

\begin{figure*}[h]
		\vspace{-0.25in}
		\begin{minipage}{0.5\textwidth}
			\centering
			\includegraphics[width=0.99\linewidth]{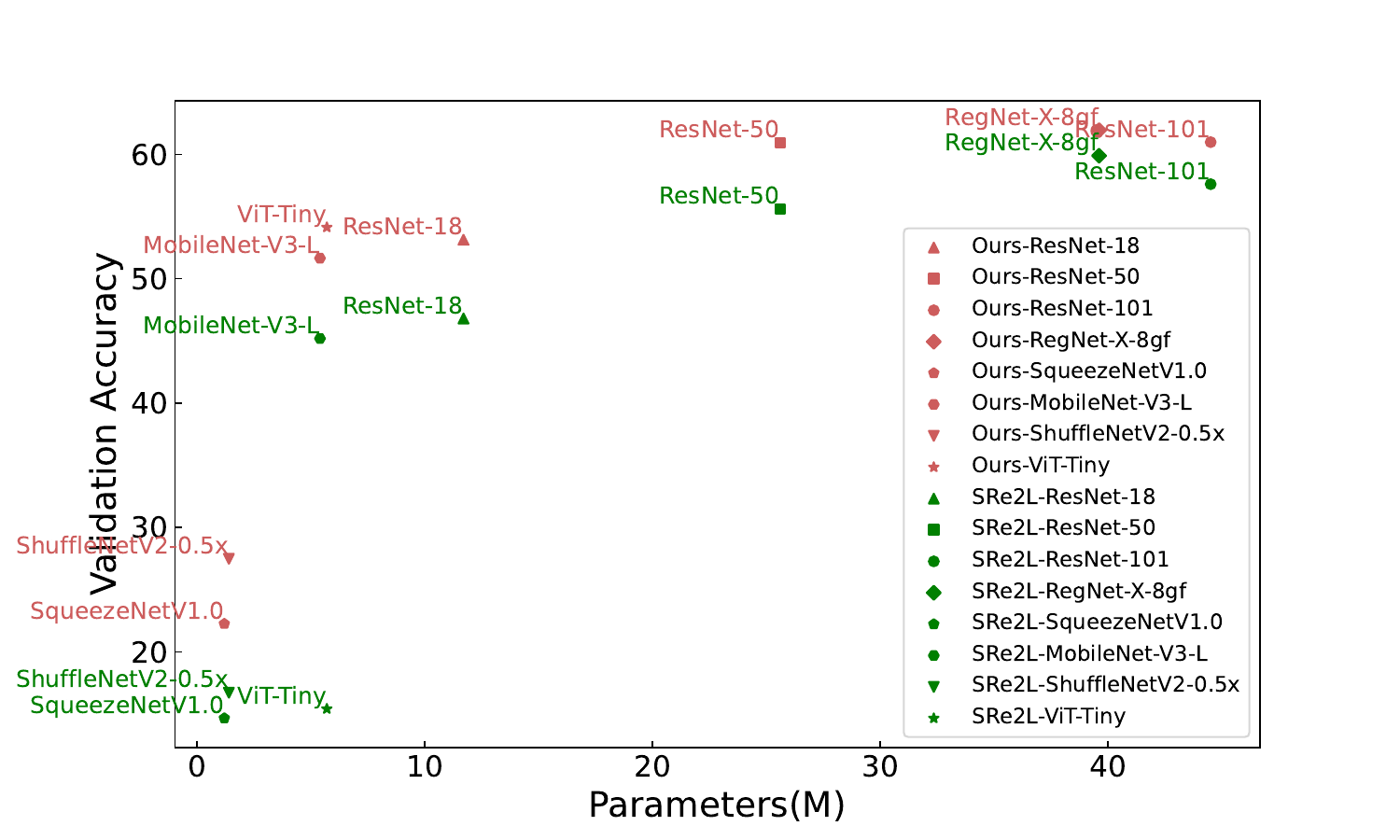}
			\vspace{-0.25in}
			\caption{Top-1 accuracy of our framework on ImageNet-1K val set using various post-training models at wide parameter scales. }
			\label{fig:cross_arch}
			\vspace{-0.15in}
		\end{minipage}
		\hfill
		\begin{minipage}{0.48\textwidth}
			\vspace{-0.2in}
			\setlength\tabcolsep{6pt}
			\centering
			\captionof{table}{Top-1 val accuracy (IPC=50) with various validation models. Recovery model is pretrained by MoCo v3.}
			\label{tab:cross_architecture}
			\resizebox{0.999\linewidth}{!}{
				\begin{tabular}{lrcc}
					\toprule
					\multicolumn{4}{c}{Various Architectures Accuracy (\%)}   \\ \midrule
					Model     &   \#Params & SRe$^2$L~\cite{yin2023squeeze}  & Ours \\
					ResNet-18 &  11.7M & 46.80 & 53.13 \\
					ResNet-50 &  25.6M & 55.60 & 60.92 \\
					ResNet-101&  44.5M & 57.60 & 60.98 \\
					RegNet-X-8gf &  39.6M & 59.89  & 61.94 \\
					SqueezeNetV1.0 &  1.2M &  14.66 & 22.25 \\
					MobileNet-V3-L &  5.4M  & 45.19 & 51.65\\
					ShuffleNetV2-0.5x & 1.4M & 16.71 & 27.48 \\
					ViT-Tiny &  5.7M &  15.41 &54.13 \\
					\bottomrule
				\end{tabular}
			} 
			\vspace{-0.15in}
		\end{minipage}

	\end{figure*}

	\noindent{\bf Cross-Architecture}. We conduct experiments to evaluate the validation accuracy of various models, using the MoCo v3 pretrained model for recovery. The models tested include ResNet-\{18, 50, 101\}, RegNet, SqueezeNet, MobileNet-v3, ShuffleNet, and Vision Transformer. The results are detailed in Table \ref{tab:cross_architecture}. Among these, RegNet-X-8gf achieves the highest accuracy.
	
	When examining the ResNet model family, we observe that increasing model size did not consistently lead to better accuracy, particularly with ResNet-50 and 101. 
	In contrast, the impressive result of RegNet-X-8gf shows the benefits of more fundamental architectural changes. Notably, our result on ViT-Tiny is significantly better than baseline SRe$^2$L, which demonstrates the effectiveness of our proposed framework. Overall, our results are consistently better than SRe$^2$L.
	
	Generally, models with larger parameter scales tend to deliver better performance across different architectures, as shown in Fig.~\ref{fig:cross_arch}. For smaller-scale models like ShuffleNet and SqueezeNet, the accuracy is limited to $\sim$20-30\%. For larger models such as ResNet, they show noticeable improvements in performance. 
	Interestingly, SRe$^2$L synthetic data on ViT-Tiny architecture does not perform well since ViT requires more training data to make it saturated, while for our framework, it dramatically improves the accuracy of SRe$^2$L by 38.72\%, demonstrating the stronger capability and potential of our proposed approach.

	\noindent{\bf Cross-Dataset}. In the first group of Table \ref{tab:cross_dataset}, we employ a pretrained MoCo v2 model, initially pretrained on ImageNet-21K, for data compression to learn the intermediate distribution of ImageNet-21K dataset, and then apply ImageNet-1K's data for classifier alignment and subsequent validation. It is observed that the accuracy achieved with this cross-dataset approach is decent, but falls short of the standard method. For a more comprehensive comparison with conventional dataset methods, we incorporate CIFAR-100 and Tiny-ImageNet into the second group of this table. 
	
	\vspace{-0.1in}
	\subsection{Application: Data-free Pruning}
	\begin{wraptable}{r}{5.0cm}
		\centering
		\vspace{-0.3in}
		\resizebox{0.95\linewidth}{!}{
			\begin{tabular}{c|cc|cc}
				\toprule
				& \multicolumn{2}{c|}{IPC 10} & \multicolumn{2}{c}{IPC 50} \\ \midrule
				Method & SRe$^2$L~\cite{yin2023squeeze}        & Ours        & SRe$^2$L~\cite{yin2023squeeze}        & Ours        \\
				Top-1  & 17.48        & \bf 23.89       & 28.19        & \bf 32.19      \\ \bottomrule  
			\end{tabular}
		}
		\vspace{-0.1in}
		\caption{Data-free pruning results.}
		\label{tab:data_free_pruning}
		\vspace{-0.35in}
	\end{wraptable}
	The condensed dataset can be effectively utilized in many aspects, serving as a valuable resource for enhancing the model's adaptability to new information and tasks over time. We apply our method to data-free pruning~\cite{srinivas2015data} similar to the protocol demonstrated in~\cite{yin2020dreaming}, using network slimming~\cite{liu2017learning} as the base pruning method and VGG-11~\cite{simonyan2014very} as the backbone. The resulting accuracy is shown in Table~\ref{tab:data_free_pruning}. It is evident that our approach outperforms the baseline approach significantly.
	
	\vspace{-0.08in}
	\section{Related Work}
	\vspace{-0.05in}
	\textbf{Dataset Distillation.} Dataset Distillation can be grouped into two categories: (1) Meta-learning based frameworks, including backpropagation-through-time methods, such as DD~\cite{wang2020dataset}, LD~\cite{bohdal2020flexible}, GTN~\cite{such2019generative}, and kernel ridge regression methods, such as KIP~\cite{nguyen2021dataset}, FRePo~\cite{zhou2022dataset}. (2) Matching based frameworks, such as Gradient Match~\cite{zhao2021dataset_DC}, Batch-Norm Match~\cite{yin2023squeeze}, Trajectory Match~\cite{cazenavette2022dataset} and Distribution Match~\cite{zhao2022dataset}. Recently, distilling on large-scale datasets has received significant attention in the community, and many works have been proposed, including~\cite{yin2023squeeze,sun2023diversity,yin2023dataset,liu2023dataset,chen2023dataset,shao2023generalized,zhou2024improve,wu2024dd,abbasi2024one,xue2024towards,qin2024distributional,gu2023efficient}. For the broader overview of related approaches for dataset distillation, please refer to~\cite{yu2023dataset,sachdeva2023data,lei2023comprehensive}. 
	
	\noindent\textbf{Self-supervised Learning.} Self-supervised learning (SSL) has emerged as a significant area of research in unsupervised representation learning. The key idea behind SSL is to use the input data itself to generate supervisory signals, often through designing pretext tasks, thus alleviating the need for large amounts of labeled data. The approaches can be divided into the following categories: (1) Contrastive based methods, such as SimCLR~\cite{chen2020simple}, MoCo~\cite{he2020momentum}, Barlow Twins~\cite{zbontar2021barlow}. (2) Clustering based methods, such has SwAV~\cite{caron2021unsupervised}, DeepCluster~\cite{caron2019deep}. (3) Distillation based methods, such as BYOL~\cite{grill2020bootstrap}, DINO~\cite{caron2021emerging}, DINO v2~\cite{oquab2023dinov2}.
	
	\vspace{-0.08in}
	\section{Conclusion}
	\vspace{-0.04in}
	In this study, we have delved into the challenges of data recovery for synthesizing by examining fluctuations in amplitude through mean and variance across channels. We pinpointed the bottlenecks of previous state-of-the-art methods that their inability to effectively scale up recovery model sizes is largely due to a flattened guidance distribution. To address this, we introduce {\em SC-DD}, a simple yet effective approach designed to amplify these fluctuations, allowing for a more nuanced capture of information during data synthesis. 
	We hope that our contributions in this work can inspire further advancements in supervised or self-supervised compression methods tailored for large-scale dataset distillation.
	
	%
	%
	\bibliographystyle{splncs04}
	\bibliography{main}

\begin{thebibliography}{10}
\providecommand{\url}[1]{\texttt{#1}}
\providecommand{\urlprefix}{URL }
\providecommand{\doi}[1]{https://doi.org/#1}

\bibitem{abbasi2024one}
Abbasi, A., Shahbazi, A., Pirsiavash, H., Kolouri, S.: One category one prompt:
  Dataset distillation using diffusion models. arXiv preprint arXiv:2403.07142
  (2024)

\bibitem{bohdal2020flexible}
Bohdal, O., Yang, Y., Hospedales, T.: Flexible dataset distillation: Learn
  labels instead of images (2020)

\bibitem{brown2020language}
Brown, T., Mann, B., Ryder, N., Subbiah, M., Kaplan, J.D., Dhariwal, P.,
  Neelakantan, A., Shyam, P., Sastry, G., Askell, A., et~al.: Language models
  are few-shot learners. Advances in neural information processing systems
  \textbf{33},  1877--1901 (2020)

\bibitem{caron2019deep}
Caron, M., Bojanowski, P., Joulin, A., Douze, M.: Deep clustering for
  unsupervised learning of visual features (2019)

\bibitem{caron2021unsupervised}
Caron, M., Misra, I., Mairal, J., Goyal, P., Bojanowski, P., Joulin, A.:
  Unsupervised learning of visual features by contrasting cluster assignments
  (2021)

\bibitem{caron2021emerging}
Caron, M., Touvron, H., Misra, I., Jégou, H., Mairal, J., Bojanowski, P.,
  Joulin, A.: Emerging properties in self-supervised vision transformers (2021)

\bibitem{cazenavette2022dataset}
Cazenavette, G., Wang, T., Torralba, A., Efros, A.A., Zhu, J.Y.: Dataset
  distillation by matching training trajectories (2022)

\bibitem{chen2023dataset}
Chen, M., Huang, B., Lu, J., Li, B., Wang, Y., Cheng, M., Wang, W.: Dataset
  distillation via adversarial prediction matching. arXiv preprint
  arXiv:2312.08912  (2023)

\bibitem{chen2020simple}
Chen, T., Kornblith, S., Norouzi, M., Hinton, G.: A simple framework for
  contrastive learning of visual representations (2020)

\bibitem{chen2020improved}
Chen, X., Fan, H., Girshick, R., He, K.: Improved baselines with momentum
  contrastive learning (2020)

\bibitem{chen2021empirical}
Chen, X., Xie, S., He, K.: An empirical study of training self-supervised
  vision transformers (2021)

\bibitem{cui2023scaling}
Cui, J., Wang, R., Si, S., Hsieh, C.J.: Scaling up dataset distillation to
  imagenet-1k with constant memory. In: International Conference on Machine
  Learning. pp. 6565--6590. PMLR (2023)

\bibitem{deng2009imagenet}
Deng, J., Dong, W., Socher, R., Li, L.J., Li, K., Fei-Fei, L.: Imagenet: A
  large-scale hierarchical image database. In: 2009 IEEE conference on computer
  vision and pattern recognition. pp. 248--255. Ieee (2009)

\bibitem{dosovitskiy2021image}
Dosovitskiy, A., Beyer, L., Kolesnikov, A., Weissenborn, D., Zhai, X.,
  Unterthiner, T., Dehghani, M., Minderer, M., Heigold, G., Gelly, S., et~al.:
  An image is worth 16x16 words: Transformers for image recognition at scale.
  In: International Conference on Learning Representations (2021)

\bibitem{grill2020bootstrap}
Grill, J.B., Strub, F., Altché, F., Tallec, C., Richemond, P.H., Buchatskaya,
  E., Doersch, C., Pires, B.A., Guo, Z.D., Azar, M.G., Piot, B., Kavukcuoglu,
  K., Munos, R., Valko, M.: Bootstrap your own latent: A new approach to
  self-supervised learning (2020)

\bibitem{gu2023efficient}
Gu, J., Vahidian, S., Kungurtsev, V., Wang, H., Jiang, W., You, Y., Chen, Y.:
  Efficient dataset distillation via minimax diffusion. arXiv preprint
  arXiv:2311.15529  (2023)

\bibitem{gulati2020conformer}
Gulati, A., Qin, J., Chiu, C.C., Parmar, N., Zhang, Y., Yu, J., Han, W., Wang,
  S., Zhang, Z., Wu, Y., et~al.: Conformer: Convolution-augmented transformer
  for speech recognition. Interspeech 2020  (2020)

\bibitem{he2020momentum}
He, K., Fan, H., Wu, Y., Xie, S., Girshick, R.: Momentum contrast for
  unsupervised visual representation learning. In: CVPR (2020)

\bibitem{he2016deep}
He, K., Zhang, X., Ren, S., Sun, J.: Deep residual learning for image
  recognition. In: Proceedings of the IEEE conference on computer vision and
  pattern recognition. pp. 770--778 (2016)

\bibitem{howard2019searching}
Howard, A., Sandler, M., Chu, G., Chen, L.C., Chen, B., Tan, M., Wang, W., Zhu,
  Y., Pang, R., Vasudevan, V., et~al.: Searching for mobilenetv3. In:
  Proceedings of the IEEE/CVF international conference on computer vision. pp.
  1314--1324 (2019)

\bibitem{hsu2021hubert}
Hsu, W.N., Bolte, B., Tsai, Y.H.H., Lakhotia, K., Salakhutdinov, R., Mohamed,
  A.: Hubert: Self-supervised speech representation learning by masked
  prediction of hidden units. IEEE/ACM Transactions on Audio, Speech, and
  Language Processing  \textbf{29},  3451--3460 (2021)

\bibitem{huang2023towards}
Huang, W., Yi, M., Zhao, X., Jiang, Z.: Towards the generalization of
  contrastive self-supervised learning. In: The Eleventh International
  Conference on Learning Representations (2023)

\bibitem{iandola2016squeezenet}
Iandola, F.N., Han, S., Moskewicz, M.W., Ashraf, K., Dally, W.J., Keutzer, K.:
  Squeezenet: Alexnet-level accuracy with 50x fewer parameters and< 0.5 mb
  model size. arXiv preprint arXiv:1602.07360  (2016)

\bibitem{jolliffe2016principal}
Jolliffe, I.T., Cadima, J.: Principal component analysis: a review and recent
  developments. Philosophical transactions of the royal society A:
  Mathematical, Physical and Engineering Sciences  \textbf{374}(2065),
  20150202 (2016)

\bibitem{kenton2019bert}
Kenton, J.D.M.W.C., Toutanova, L.K.: Bert: Pre-training of deep bidirectional
  transformers for language understanding. In: Proceedings of NAACL-HLT. pp.
  4171--4186 (2019)

\bibitem{kirichenko2022last}
Kirichenko, P., Izmailov, P., Wilson, A.G.: Last layer re-training is
  sufficient for robustness to spurious correlations. In: The Eleventh
  International Conference on Learning Representations (2022)

\bibitem{krizhevsky2009learning}
Krizhevsky, A., Hinton, G., et~al.: Learning multiple layers of features from
  tiny images  (2009)

\bibitem{krizhevsky2012imagenet}
Krizhevsky, A., Sutskever, I., Hinton, G.E.: Imagenet classification with deep
  convolutional neural networks. Advances in neural information processing
  systems  \textbf{25} (2012)

\bibitem{kumar2022fine}
Kumar, A., Raghunathan, A., Jones, R., Ma, T., Liang, P.: Fine-tuning can
  distort pretrained features and underperform out-of-distribution. In:
  International Conference on Learning Representations (2022)

\bibitem{le2015tiny}
Le, Y., Yang, X.: Tiny imagenet visual recognition challenge. CS 231N
  \textbf{7}(7), ~3 (2015)

\bibitem{lee2021predicting}
Lee, J.D., Lei, Q., Saunshi, N., Zhuo, J.: Predicting what you already know
  helps: Provable self-supervised learning. Advances in Neural Information
  Processing Systems  \textbf{34},  309--323 (2021)

\bibitem{lei2023comprehensive}
Lei, S., Tao, D.: A comprehensive survey to dataset distillation. arXiv
  preprint arXiv:2301.05603  (2023)

\bibitem{liu2023dataset}
Liu, H., Xing, T., Li, L., Dalal, V., He, J., Wang, H.: Dataset distillation
  via the wasserstein metric. arXiv preprint arXiv:2311.18531  (2023)

\bibitem{liu2017learning}
Liu, Z., Li, J., Shen, Z., Huang, G., Yan, S., Zhang, C.: Learning efficient
  convolutional networks through network slimming. In: Proceedings of the IEEE
  international conference on computer vision. pp. 2736--2744 (2017)

\bibitem{ma2018shufflenet}
Ma, N., Zhang, X., Zheng, H.T., Sun, J.: Shufflenet v2: Practical guidelines
  for efficient cnn architecture design. In: Proceedings of the European
  conference on computer vision (ECCV). pp. 116--131 (2018)

\bibitem{macqueen1967some}
MacQueen, J., et~al.: Some methods for classification and analysis of
  multivariate observations. In: Proceedings of the fifth Berkeley symposium on
  mathematical statistics and probability. vol.~1, pp. 281--297. Oakland, CA,
  USA (1967)

\bibitem{nguyen2021dataset}
Nguyen, T., Chen, Z., Lee, J.: Dataset meta-learning from kernel
  ridge-regression (2021)

\bibitem{openai2023gpt4}
OpenAI: Gpt-4 technical report (2023)

\bibitem{oquab2023dinov2}
Oquab, M., Darcet, T., Moutakanni, T., Vo, H., Szafraniec, M., Khalidov, V.,
  Fernandez, P., Haziza, D., Massa, F., El-Nouby, A., Assran, M., Ballas, N.,
  Galuba, W., Howes, R., Huang, P.Y., Li, S.W., Misra, I., Rabbat, M., Sharma,
  V., Synnaeve, G., Xu, H., Jegou, H., Mairal, J., Labatut, P., Joulin, A.,
  Bojanowski, P.: Dinov2: Learning robust visual features without supervision
  (2023)

\bibitem{qin2024distributional}
Qin, T., Deng, Z., Alvarez-Melis, D.: Distributional dataset distillation with
  subtask decomposition. arXiv preprint arXiv:2403.00999  (2024)

\bibitem{radosavovic2020designing}
Radosavovic, I., Kosaraju, R.P., Girshick, R., He, K., Doll{\'a}r, P.:
  Designing network design spaces. In: Proceedings of the IEEE/CVF conference
  on computer vision and pattern recognition. pp. 10428--10436 (2020)

\bibitem{rosenblatt1956central}
Rosenblatt, M.: A central limit theorem and a strong mixing condition.
  Proceedings of the national Academy of Sciences  \textbf{42}(1),  43--47
  (1956)

\bibitem{sachdeva2023data}
Sachdeva, N., McAuley, J.: Data distillation: A survey. arXiv preprint
  arXiv:2301.04272  (2023)

\bibitem{shao2023generalized}
Shao, S., Yin, Z., Zhou, M., Zhang, X., Shen, Z.: Generalized large-scale data
  condensation via various backbone and statistical matching. arXiv preprint
  arXiv:2311.17950  (2023)

\bibitem{shen2021partial}
Shen, Z., Liu, Z., Qin, J., Savvides, M., Cheng, K.T.: Partial is better than
  all: revisiting fine-tuning strategy for few-shot learning. In: Proceedings
  of the AAAI Conference on Artificial Intelligence. vol.~35, pp. 9594--9602
  (2021)

\bibitem{shen2022fast}
Shen, Z., Xing, E.: A fast knowledge distillation framework for visual
  recognition. In: European Conference on Computer Vision. pp. 673--690.
  Springer (2022)

\bibitem{simonyan2014very}
Simonyan, K., Zisserman, A.: Very deep convolutional networks for large-scale
  image recognition. arXiv preprint arXiv:1409.1556  (2014)

\bibitem{srinivas2015data}
Srinivas, S., Babu, R.V.: Data-free parameter pruning for deep neural networks.
  arXiv preprint arXiv:1507.06149  (2015)

\bibitem{such2019generative}
Such, F.P., Rawal, A., Lehman, J., Stanley, K.O., Clune, J.: Generative
  teaching networks: Accelerating neural architecture search by learning to
  generate synthetic training data (2019)

\bibitem{sun2023diversity}
Sun, P., Shi, B., Yu, D., Lin, T.: On the diversity and realism of distilled
  dataset: An efficient dataset distillation paradigm. arXiv preprint
  arXiv:2312.03526  (2023)

\bibitem{wang2022cafe}
Wang, K., Zhao, B., Peng, X., Zhu, Z., Yang, S., Wang, S., Huang, G., Bilen,
  H., Wang, X., You, Y.: Cafe: Learning to condense dataset by aligning
  features (2022)

\bibitem{wang2020dataset}
Wang, T., Zhu, J.Y., Torralba, A., Efros, A.A.: Dataset distillation (2020)

\bibitem{wu2024dd}
Wu, Y., Du, J., Liu, P., Lin, Y., Cheng, W., Xu, W.: Dd-robustbench: An
  adversarial robustness benchmark for dataset distillation. arXiv preprint
  arXiv:2403.13322  (2024)

\bibitem{xue2024towards}
Xue, E., Li, Y., Liu, H., Shen, Y., Wang, H.: Towards adversarially robust
  dataset distillation by curvature regularization. arXiv preprint
  arXiv:2403.10045  (2024)

\bibitem{yin2020dreaming}
Yin, H., Molchanov, P., Alvarez, J.M., Li, Z., Mallya, A., Hoiem, D., Jha,
  N.K., Kautz, J.: Dreaming to distill: Data-free knowledge transfer via
  deepinversion. In: Proceedings of the IEEE/CVF Conference on Computer Vision
  and Pattern Recognition. pp. 8715--8724 (2020)

\bibitem{yin2023dataset}
Yin, Z., Shen, Z.: Dataset distillation in large data era. arXiv preprint
  arXiv:2311.18838  (2023)

\bibitem{yin2023squeeze}
Yin, Z., Xing, E., Shen, Z.: Squeeze, recover and relabel: Dataset condensation
  at imagenet scale from a new perspective. In: NeurIPS (2023)

\bibitem{yosinski2014transferable}
Yosinski, J., Clune, J., Bengio, Y., Lipson, H.: How transferable are features
  in deep neural networks? Advances in neural information processing systems
  \textbf{27} (2014)

\bibitem{yu2023dataset}
Yu, R., Liu, S., Wang, X.: Dataset distillation: A comprehensive review. arXiv
  preprint arXiv:2301.07014  (2023)

\bibitem{zbontar2021barlow}
Zbontar, J., Jing, L., Misra, I., LeCun, Y., Deny, S.: Barlow twins:
  Self-supervised learning via redundancy reduction (2021)

\bibitem{zhao2022dataset}
Zhao, B., Bilen, H.: Dataset condensation with distribution matching (2022)

\bibitem{zhao2021dataset_DC}
Zhao, B., Mopuri, K.R., Bilen, H.: Dataset condensation with gradient matching
  (2021)

\bibitem{zhou2024improve}
Zhou, B., Zhong, L., Chen, W.: Improve cross-architecture generalization on
  dataset distillation. arXiv preprint arXiv:2402.13007  (2024)

\bibitem{zhou2022dataset}
Zhou, Y., Nezhadarya, E., Ba, J.: Dataset distillation using neural feature
  regression (2022)

\end{thebibliography}
	
	\newpage
	
	\appendix
	
	\section*{\Large{Appendix}}
	
	In the appendix, we provide more details for supplementing the main paper, including:
	\vspace{0.05in}
	
	\noindent{• Section~\ref{proof}: Proofs for Section 2 in the main paper.}
	
	\noindent{• Section~\ref{computation}: Training time/computational cost analysis.}
	
	\noindent{• Section~\ref{details}: More implementation details.}
	
	\noindent{• Section~\ref{pretrain}: Accuracy of self-supervised pretrained models for image synthesis/recovery.}
	
	\noindent{• Section~\ref{additional_res}: Additional ablation studies.}
	
	\noindent{• Section~\ref{additional_vis}: Additional visualization.}
	
	\section{Proofs for Section 2} \label{proof}
	
	\noindent{\bf Restatement of Equation 7 in the main paper.} Assuming that the amount of pre-training data is large, and the data is diverse enough, according to the central limit theorem~\cite{rosenblatt1956central}, the features follow a Gaussian distribution post-normalization, we can simplify the entropy of both distributions to:
	\vspace{-0.08in}
	\begin{equation}
		H(\Theta_{ssl}) = \frac{1}{2} \log(2 \pi e \sigma_{ssl}^2), H(\Theta_{sl}) = \frac{1}{2} \log(2 \pi e \sigma_{sl}^2)
	\end{equation}
	\noindent{\bf \em Proof:} The entropy of a continuous random variable (feature representation) is a measure of the uncertainty associated with its possible outcomes. For a Gaussian distribution, which is a continuous distribution, the entropy can be calculated using the differential entropy formula for a normal distribution. The probability density function (PDF) for a Gaussian distribution is given by:
	\begin{equation}
		p(z) = \frac{1}{\sqrt{2\pi\sigma^2}} \exp\left(-\frac{(z-\mu)^2}{2\sigma^2}\right) 
	\end{equation} 
	where \( z \) is the input feature representation, \( \mu \) is the mean of the feature representation distribution, \( \sigma \) is the standard deviation of the distribution, \( \sigma^2 \) is the variance.
	
	The differential entropy \( H \) for a continuous random representation with probability density function \( p(z) \) is given by:
	\begin{equation}
		H(\Theta) = -\int_{-\infty}^{\infty} p(z) \log(p(z)) \,dz   
	\end{equation}
	
	Substituting the PDF of the Gaussian distribution into the entropy formula, we get:
	\begin{equation}
		\begin{aligned}
			H(\Theta) =& -\int_{-\infty}^{\infty} \frac{1}{\sqrt{2\pi\sigma^2}} \exp\left(-\frac{(z-\mu)^2}{2\sigma^2}\right) \\&\log\left(\frac{1}{\sqrt{2\pi\sigma^2}} \exp\left(-\frac{(z-\mu)^2}{2\sigma^2}\right)\right) \,dz 
		\end{aligned}
	\end{equation}
	
	Then, we split the logarithm into two parts using the logarithm property: 
	\begin{equation}
		\begin{aligned}
			H(\Theta) = & -\int_{-\infty}^{\infty} p(z) [\log\left(\frac{1}{\sqrt{2\pi\sigma^2}}\right) \\&+ \log\left(\exp\left(-\frac{(z-\mu)^2}{2\sigma^2}\right)\right)] \,dz 
		\end{aligned}
	\end{equation}
	
	We further simplify the second term inside the integral by using the fact that \( \log(\exp(a)) = a \):
	\begin{equation}
		H(\Theta) = -\int_{-\infty}^{\infty} p(z) \left[-\log(\sqrt{2\pi\sigma^2}) - \frac{(z-\mu)^2}{2\sigma^2}\right] \,dz  
	\end{equation}
	
	The integral of the Gaussian PDF \( p(z) \) over its entire range is 1, and the integral of \( p(z) \) times a quadratic function centered on its mean is simply the variance \( \sigma^2 \). So, the entropy simplifies to:
	\begin{equation}
		\begin{aligned}
			H(\Theta) & =-\int_{-\infty}^{\infty} p(z) \log p(z) \mathrm{d} z \\
			& =-\mathbb{E}\left[\log \left[\left(2 \pi \sigma^2\right)^{-1 / 2} \exp \left(-\frac{1}{2 \sigma^2}(z-\mu)^2\right)\right]\right] \\
			& =\frac{1}{2} \log \left(2 \pi \sigma^2\right)+\frac{1}{2 \sigma^2} \mathbb{E}\left[(z-\mu)^2\right] \\
			& =\frac{1}{2} \log \left(2 \pi \sigma^2\right)+\frac{1}{2} .
		\end{aligned}   
	\end{equation}
	where $\mathbb{E}\left[(z-\mu)^2\right]=\sigma^2$. Finally, the entropy of $z$ is a function of its variance $\sigma^2$. 
	\begin{equation}
		H(\Theta) = \frac{1}{2} \log(2\pi\sigma^2) + \frac{1}{2}   
	\end{equation}
	Since \( e \) is the base of the natural logarithm, the \( \frac{1}{2} \) outside the log can be taken inside to give \( e \):
	\begin{equation}
		H(\Theta) = \frac{1}{2} \log(2\pi e \sigma^2) 
	\end{equation}
	This final expression gives us the entropy of a Gaussian distribution in terms of its variance \( \sigma^2 \). The entropy is maximized when the variance is large, indicating that a broader distribution (more uncertainty) leads to higher entropy.
	
	\begin{figure*}[t]
		\centering
		\includegraphics[width=0.7\linewidth]{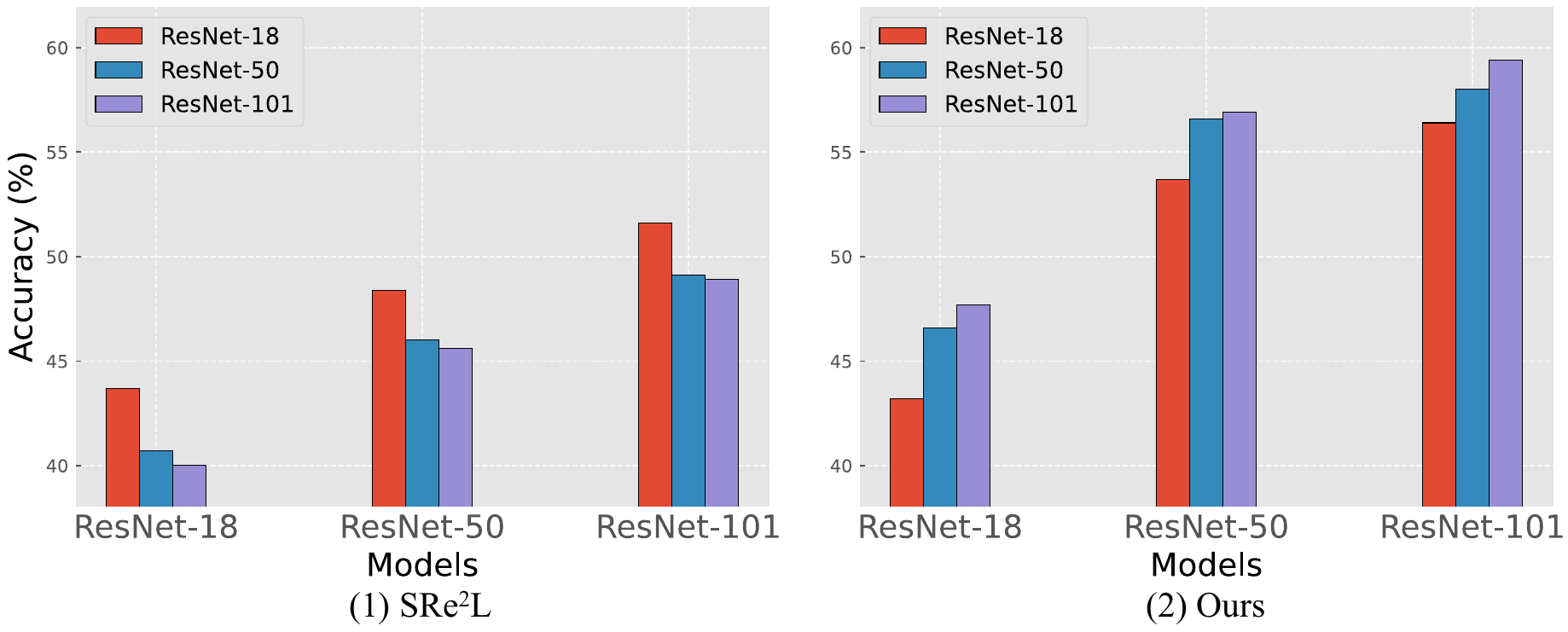}
		\vspace{-0.16in}
		\caption{Top-1 accuracy on distilled images from different recovery model architectures with various validation models. In each subfigure, x-axis represents the validation model trained on the images synthesized from ResNet-\{18, 50, 101\} in each group of the subfigures, i.e., each column in the histogram.}
		\label{fig:ablation_arch}
		\vspace{-0.1in}
	\end{figure*}
	
	\section{Computational Cost Analysis} \label{computation}
	
	We present the consumption of GPU hours in the processes of self-supervised squeezing, image synthesis/recovery, and validation/post-training.
	
	\noindent{\textbf{Squeezing:}} The time costs associated with self-supervised pre-training and linear probing are listed under various model architectures, as outlined in Table \ref{tab:GPUTime_pretrain}.
	
	\noindent{\textbf{Recovery:}} The time costs incurred during the recovery process, considering different model architectures (ResNet-18, ResNet-50, and ResNet-101), are detailed in Table \ref{tab:GPUTime_recover}. 
	
	\noindent{\textbf{Validation:}} The time costs in the validation phase, subject to varying IPC, are presented in Table \ref{tab:GPUTime_post}. An increase in IPC is associated with an augmentation in time costs, concomitant with an increase in validation accuracy. 
	
	\begin{table}[h]
\vspace{-0.06in}
\centering
\resizebox{0.6\linewidth}{!}{
\begin{tabular}{@{}cccc@{}}
\toprule
\multirow{2}{*}{Method}  & \multirow{2}{*}{Model} & \multicolumn{2}{c}{Time (hours/per epoch)} \\ \cmidrule(l){3-4} 
                         &                        & pre-training   & linear probing  \\ \midrule
\multirow{3}{*}{MoCo v2} & ResNet-18              & 0.14          & 0.04            \\
                         & ResNet-50              & 0.22           & 0.06            \\
                         & ResNet-101             & 0.26           & 0.13           \\ \midrule
MoCo v3                  & ResNet-50              & --           & 0.06            \\ \midrule
\multirow{4}{*}{SwAV}    & ResNet-50              & --             & 0.06            \\
                         & ResNet-50-w2           & --             & 0.09                \\
                         & ResNet-50-w4           & --             & 0.23            \\
                         & ResNet-50-w5           & --             & 0.33            \\ \midrule
DINO                     & ResNet-50              & --             & 0.06            \\ \bottomrule
\end{tabular}
}
\caption{Time consumption per epoch of self-supervised pre-training on ImageNet-1K with 4 $\times$ NVIDIA A100 (40G) GPUs. Pretraining time estimations for MoCo v3, SwAV and DINO are omitted as they are entirely the same as the default pretraning overhead. We use the off-the-shelf official pretrained models for them, and train a linear probing layer for each by ourselves.}
\label{tab:GPUTime_pretrain}
\end{table}

	\begin{table}[h]
\centering
\begin{tabular}{@{}cccc@{}}
\toprule
Recover Architecture & GPU hours \\ \midrule
ResNet18        &   7.78 \\
ResNet50        &  25.04  \\
ResNet101       &  35.44  \\
\bottomrule
\end{tabular}
\caption{Time consumption to generate distilled images on ImageNet-1K with IPC 50 using one 4090 GPU.}
\label{tab:GPUTime_recover}
\vspace{-0.2in}
\end{table}

	\begin{table}[h]
\centering
\begin{tabular}{@{}cc@{}}
\toprule
 IPC & GPU hours \\ \midrule
10        &   0.75 \\
50        &  2.53  \\
100       & 4.58 \\
\bottomrule
\end{tabular}
\caption{Post training GPU hours for every 100 epoch with MoCo v3 ResNet50 under various recovery IPCs using one 4090 GPU.}
\label{tab:GPUTime_post}
\vspace{-0.2in}
\end{table}

	\section{Implementation Details} \label{details}
	
	We outline the parameter configurations for the processes of squeezing, recovering, and validation across CIFAR-100, Tiny-ImageNet, and ImageNet, as detailed in Table~\ref{tab:config-cifar-mix}, Table~\ref{tab:config-tiny-mix}, and Table~\ref{tab:config-IM-mix}, respectively.
	
	\noindent{\textbf{CIFAR-100.}} The MoCo\footnote{\url{https://colab.research.google.com/github/facebookresearch/moco/blob/colab-notebook/colab/moco_cifar10_demo.ipynb}.} framework is used for pretraining ResNet-{18, 50} backbones, achieving the linear accuracy presented in the first group of Table~\ref{cifar_tiny}. Hyper-parameters from the second column of Table~\ref{tab:config-cifar-mix} are utilized for the pretraining. Subsequently, the parameters from the third and fourth columns in the same table are employed, leading to a validation accuracy of 58.7\% under IPC 50.
	
	\noindent{\textbf{Tiny-ImageNet.}} The same MoCo framework as used on CIFAR-100 is adapted to pretrain ResNet-{18, 50} backbones on Tiny-ImageNet dataset, their linear accuracy is provided in the second group of Table~\ref{cifar_tiny}. The parameter settings outlined in Table~\ref{tab:config-tiny-mix} contribute to achieving the validation accuracy reported in the main paper.
	
	\noindent{\textbf{ImageNet-1K.}} The linear probing of MoCo v3 is employed using parameters specified in the second column of Table~\ref{tab:config-IM-mix}. Subsequently, parameters from the third and fourth columns in the same table are employed for the recovery and validation phases. Notably, a 600-ep training budget in validation phase is performed for achieving the final performance in the main paper, while a 300-ep training budget is applied in all other experiments for ablation studies.
	
	\noindent{Additionally}, in Table \ref{tab:loss-parameter}, we provide the details of self-supervised objectives for different models, including MoCo v2, MoCo v3, SwAV, and DINO. The table also encompasses the corresponding hyper-parameters of learning rate and Batch Normalization coefficient in recovery phase.
	
	\begin{table}[h]
\vspace{-0.1in}
\centering
\resizebox{0.98\linewidth}{!}{
\begin{tabular}{l|c|c|c|c}
\toprule
pretraining type    & method & loss objective & lr   & BN coefficient \\ \midrule
\multirow{2}{*}{contrastive} & MoCo v2     &  \multirow{2}{*}{$\mathcal{L}=-\log \frac{\exp \left(q \cdot k_{+} / \tau\right)}{\sum_{i=0}^K \exp \left(q \cdot k_i / \tau\right)}$}    & 0.35 & 0.25           \\ 
                    & MoCo v3     &                & 0.25 & 0.0005         \\ \hline
clustering          & SwAV   &    $\mathcal{L}=-\sum_k \mathbf{q}_s^{(k)} \log \mathbf{p}_t^{(k)}-\sum_k \mathbf{q}_t^{(k)} \log \mathbf{p}_s^{(k)}$   & 0.3  & 0.001          \\ \hline
distillation        & DINO   &   $\min _{\theta_s} H\left(P_t(x), P_s(x)\right)$    & 0.25 & 0.01           \\ 
\bottomrule
\end{tabular}
} 
\caption{{Loss objectives and their corresponding hyperparameters for different pertaining methods.}}
\label{tab:loss-parameter}
\end{table}

	\begin{table}[]
\centering
\resizebox{0.6\linewidth}{!}{
\begin{tabular}{@{}c|ccc@{}}
Config             & Pretrain  & Recover  & Validation    \\ \midrule
$\alpha_{BN}$      &   -       & 0.005    & - \\
optimizer          &   SGD     & Adam     & AdamW      \\
base learning rate &   0.06    & 0.4      & {0.005} \\
weight decay       &   5e-4    &  1e-4  & {0.01} \\
optimizer momentum &  0.9     & (0.5, 0.9)   & (0.9, 0.999) \\
batch size         & 512       & 100      & {64} \\
learning rate schedule & cosine & cosine & cosine \\
recovering iteration & -       & 1,000     & - \\
training epoch     & {200}        & -        & {200} \\
\end{tabular}
}
\caption{Hyper-parameter setting on CIFAR-100. Optimizer parameters ($\beta_1, \beta_2$) represent the exponential decay rate for the first and second moment estimates.}
\label{tab:config-cifar-mix}
\end{table}

	\begin{table}[t]
\centering
\resizebox{0.6\linewidth}{!}{
\begin{tabular}{@{}c|ccc@{}}
Config             & Pretrain  & Recover  &  Validation   \\ \midrule
$\alpha_{BN}$      &   -       &  0.1     &  -  \\
optimizer          &   SGD     &  Adam    & SGD   \\
base learning rate &   0.06    &  0.6     & 0.2 \\
weight decay       &   5e-4    &  1e-4       & 1e-4 \\
optimizer momentum &  0.9     &  (0.5, 0.9)    & 0.9 \\
batch size         & 512       &  100     & 64 \\
learning rate schedule & cosine & cosine & cosine \\
recovering iteration & -       & 1,000     & - \\
augmentation       &  -        & RRC*  & {RandAugment} \\
training epoch     & {200}         & -        & {500} \\
\end{tabular}
}
\caption{Hyperparameter setting on Tiny-ImageNet. * represents {\em RandomResizedCrop}. Optimizer parameters ($\beta_1, \beta_2$) represent the exponential decay rate for the first and second moment estimates. We choose a slightly larger post-training budget, which aligns with RDED~\cite{sun2023diversity}.} 
\label{tab:config-tiny-mix}
\vspace{-0.2in}
\end{table}

	\begin{table}[t]
\centering
\resizebox{0.6\linewidth}{!}{
\begin{tabular}{@{}c|ccc@{}}
Config             & Linear probing & Recover & Validation    \\ 
\midrule
$\alpha_{BN}$      &  -       &  0.0005 & -          \\
optimizer          &  SGD     &  Adam  & AdamW   \\
base learning rate &  0.03    & 0.25    &  0.001  \\
weight decay       &  {1e-4}   &  1e-4      &   0.01 \\
optimizer momentum &  0.9     & (0.5, 0.9) &(0.9, 0.999)  \\
batch size         &  {256}     &  50     &  64  \\
learning rate schedule & cosine  &  cosine & cosine\\
recovering iteration & -  &  1,000  & - \\
augmentation       &   - &   RRC* & RRC* \\
training epoch     &   {200}     &  -       & 300 \\
\end{tabular}
}
\caption{Hyper-parameter setting on ImageNet-1K. * represents {\em RandomResizedCrop}. Optimizer parameters ($\beta_1, \beta_2$) represent the exponential decay rate for the first and second moment estimates.}
\label{tab:config-IM-mix}
\vspace{-0.2in}
\end{table}
	
	\section{Accuracy of Self-supervised Models for Recovery} \label{pretrain}
	
	We provide all our linear probing models that we used for image synthesis/recovery. On CIFAR-100 and Tiny-ImageNet datasets, we utilize ResNet-18 and ResNet50 models that are pretrained using MoCo~\cite{he2020momentum}. The results are provided in Table~\ref{cifar_tiny}. On ImageNet-1K, we employ various model architectures pretrained on MoCo V2~\cite{chen2020improved}, v3~\cite{chen2021empirical}, SwAV~\cite{caron2021unsupervised} and DINO~\cite{caron2021emerging} with different pertaining budgets. The detailed results are shown in Table~\ref{tab:lincls_IN}.
	
	\begin{table}[t]
\centering
\resizebox{0.55\linewidth}{!}{
\begin{tabular}{@{}cccc@{}}
\toprule
method                   & model                      & pretrain epochs & linear acc. \\ \midrule
\multirow{8}{*}{MoCo v2} & \multirow{4}{*}{ResNet-50} & 100             & 64.7     \\
                         &                            & 200             & 67.6        \\
                         &                            & 400             & 69.6       \\
                         &                            & 800             & 71.1        \\ \cmidrule(l){2-4} 
                         & ResNet-18                  & 200             & 53.3       \\
                         & ResNet-18                  & 800             & 53.5       \\
                         & ResNet-101                 & 200             & 69.9    \\
                         & ResNet-101                 & 800             & 73.0    \\ \midrule
\multirow{3}{*}{MoCo v3} & \multirow{3}{*}{ResNet-50} & 100             & 68.9    \\ 
                         &                            & 300             & 72.8    \\
                         &                            & 1,000           & 74.6    \\ \midrule  
\multirow{7}{*}{SwAV}    & \multirow{4}{*}{ResNet-50} & 100             & 72.0        \\
                         &                            & 200             & 73.8        \\
                         &                            & 400             & 74.5        \\
                         &                            & 800             & 75.3    \\ \cmidrule(l){2-4}  
                         & ResNet-50-w2               & 400             & 77.2        \\
                         & ResNet-50-w4               & 400             & 77.4        \\
                         & ResNet-50-w5               & 400             & 77.9        \\ \midrule
\multirow{2}{*}{DINO}    & \multirow{2}{*}{ResNet-50} & 100             & 66.8      \\
                         &                            & 800             & 75.3        \\ \bottomrule   
\end{tabular}
}
\caption{Linear classification results on self-supervised models. } 
\label{tab:lincls_IN}
\vspace{-0.15in}
\end{table}

\begin{table}[h]
\centering
\resizebox{0.39\linewidth}{!}{
\begin{tabular}{@{}ccc@{}}
\toprule
dataset                        & model     & linear acc \\ \midrule
\multirow{2}{*}{CIFAR-100}     & ResNet-18 & 60.8       \\
                               & ResNet-50 & 60.8       \\ \midrule
\multirow{2}{*}{Tiny-ImageNet} & ResNet-18 & 45.4       \\
                               & ResNet-50 & 51.0       \\ \bottomrule
\end{tabular}
}
\caption{Linear classification results on small datasets, using MoCo to pretrain the backbone for 200 epochs.}
\label{cifar_tiny}
\vspace{-0.1in}
\end{table}

	\section{Additional Ablation Studies} \label{additional_res}
	
	We examine two factors that could influence the final accuracy in validation, including training budget and batch size.
	
	\noindent{\bf Post-training budget.} To investigate the impact of epochs on validation accuracy, we conduct experiments by evaluating at 10 distinct budgets of epochs (from 100 to 1,000) using distilled images with IPC 10 and 50, as illustrated in Fig.~\ref{fig:training_budget}. The observed gradual improvement in accuracy with increasing epochs supports the assertion that augmenting the training epoch contributes to enhanced accuracy. However, this correlation holds true within a specific epoch range. As epochs continue to increase, the rate of validation accuracy improvement diminishes. Notably, an augmentation of epochs from 900 to 1,000 for IPC of 10 resulted in a decline in validation accuracy from 54.59\% to 54.45\%. 
	
	\begin{figure}[h]
		\centering
		\includegraphics[width=0.8\linewidth]{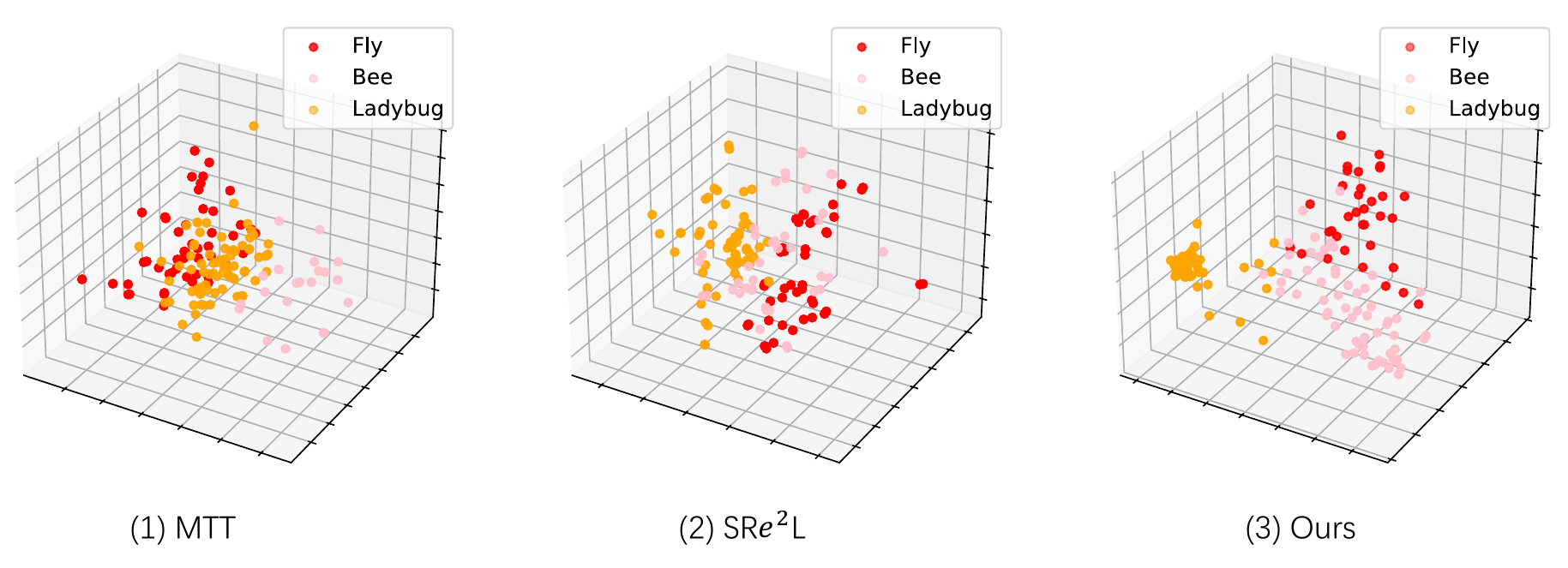}
		\vspace{-0.1in}
		\caption{Clustering visualization of semantically-similar synthetic data by MTT, SRe$^2$L and ours with three similar classes: Fly, Bee, and Ladybug.}
		\label{fig:kmeans_similar}
		\vspace{-0.1in}
	\end{figure}
	
	\noindent{\bf Batch size.} 
	In Table~\ref{tab:batch_size}, we provide the Top-1 accuracy of various batch sizes in the validation phase on the ImageNet-1K dataset. The accuracy demonstrates an upward trend as the batch size decreases, reaching its peak at a batch size of 64. Subsequently, a reduction in accuracy is observed with further decreases in batch size. Notably, the optimal accuracy is attained when employing a batch size of 64, prompting its consistent utilization in all our experiments conducted on ImageNet-1K.
	
	Moreover, for a more comprehensive understanding of our method's capability to the state-of-the-art SRe$^2$L, we illustrate Fig.~\ref{fig:ablation_arch}, which shows that, contrary to the observed trend in SRe$^2$L method (left subfigure) where larger models used for recovery lead to decreased post-validation performance, our method exhibits an inverse relationship. As the size of the recovery models increases, performance improves, indicating a higher potential in our approach.
	
	\begin{table}[h]
\centering
\begin{tabular}{@{}cccc@{}}
\toprule
Batch Size & Accuracy \\ \midrule
128        & 58.68   \\
96         & 59.34   \\
48         & 59.98   \\
64         & {\bf 60.92}   \\
32         & 58.31   \\ \bottomrule
\end{tabular}
\caption{Validation results on synthetic dataset recovered from MoCo v3 300-ep pretrained ResNet-50 with various batch sizes.}
\label{tab:batch_size}
\vspace{-0.1in}
\end{table}

	\begin{table}[h]
\centering
\resizebox{0.65\linewidth}{!}{
\begin{tabular}{@{}ccccc@{}}
\toprule
\multicolumn{2}{c}{\multirow{2}{*}{recovery model}} & \multicolumn{3}{c}{validation accuracy (\%)} \\ \cline{3-5} 
\multicolumn{2}{c}{}                   & ResNet-18 & ResNet-101 & RegNet-X-8gf \\

\midrule
\multirow{3}{*}{\rotatebox{90}{Ours}}  & ResNet-18  &  43.15 & 53.72 & 56.39 \\
 & ResNet-50  & 46.62 & 56.60   &  58.00  \\
 & RegNet-101 &  47.71  & 56.88   & 59.42  \\ 

\midrule
\multicolumn{2}{c}{\multirow{2}{*}{recovery model}} & \multicolumn{3}{c}{validation accuracy (\%)} \\ \cline{3-5} 
\multicolumn{2}{c}{}                   & ResNet-18 & ResNet-50 & ResNet-101 \\ 

\midrule
\multirow{3}{*}{\rotatebox{90}{SRe$^2$L}}  & ResNet-18    & 43.69  & 48.36   & 51.57  \\
 & ResNet-50 &  40.66 & 46.02  & 49.06 \\
 & ResNet-101 &  39.95 & 45.56   &  48.92   \\

\bottomrule
\end{tabular}
}
\caption{Top-1 accuracy of ours (MoCo v2 based) and SRe$^2$L on ImageNet-1K dataset.}
\label{tab:fig_2}
\vspace{-0.1in}
\end{table}
	
	\section{Additional Visualization} \label{additional_vis}
	
	We provide additional clustering visualization of semantically similar classes by MTT, SRe$^2$L, and our proposed SC-DD with three classes: Fly, Bee, and Ladybug. As shown in Fig.~\ref{fig:kmeans_similar},  it can be observed that the data generated by our approach is more distinguishable in the low dimension space, indicating that thay have learned more semantic information in the synthetic data.
	
\begin{figure*}[h]
		\centering
		\includegraphics[width=0.9\linewidth]{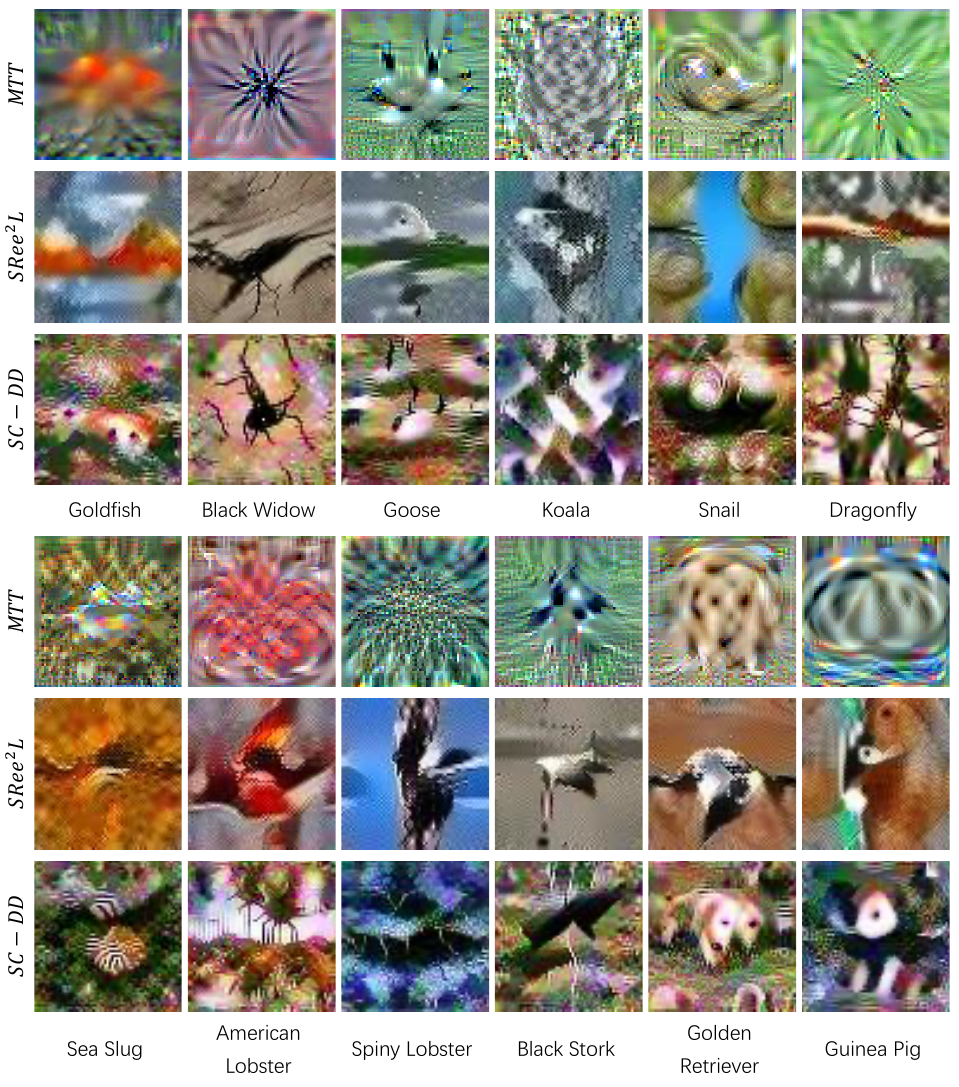}
		\vspace{-0.1in}
		\caption{Comparative synthetic data on Tiny-ImageNet from MTT, SRe$^2$L and our {\em SC-DD}.}
		\label{fig:tiny_contrast}
	\end{figure*}

\begin{figure*}[h]
		\centering
		\includegraphics[width=0.9\linewidth]{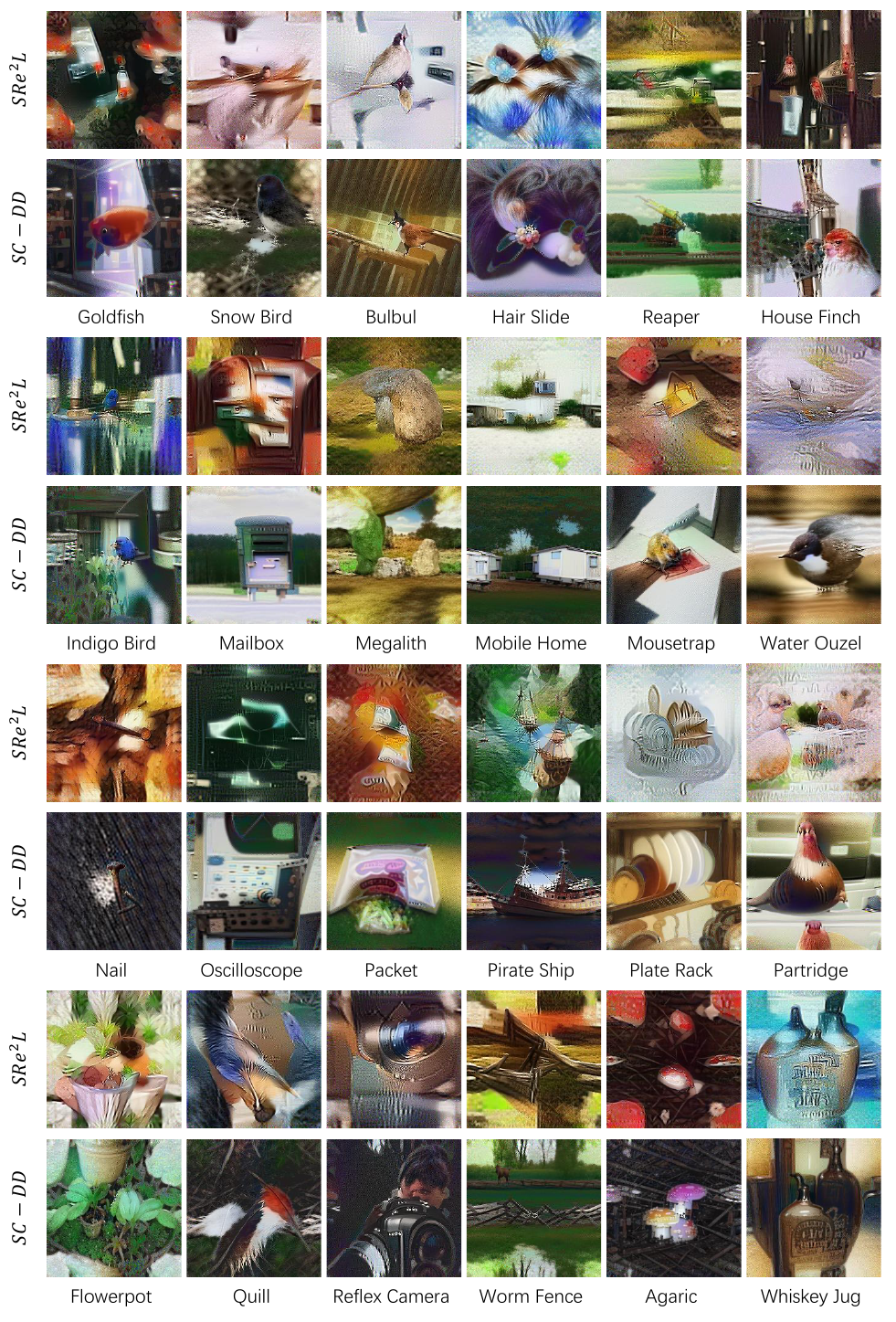}
		\vspace{-0.1in}
		\caption{Comparative synthetic data on ImageNet-1K from SRe$^2$L and our {\em SC-DD}.}
		\label{fig:IM_contrast}
	\end{figure*}

\begin{figure*}[h]
		\centering
		\includegraphics[width=0.9\linewidth]{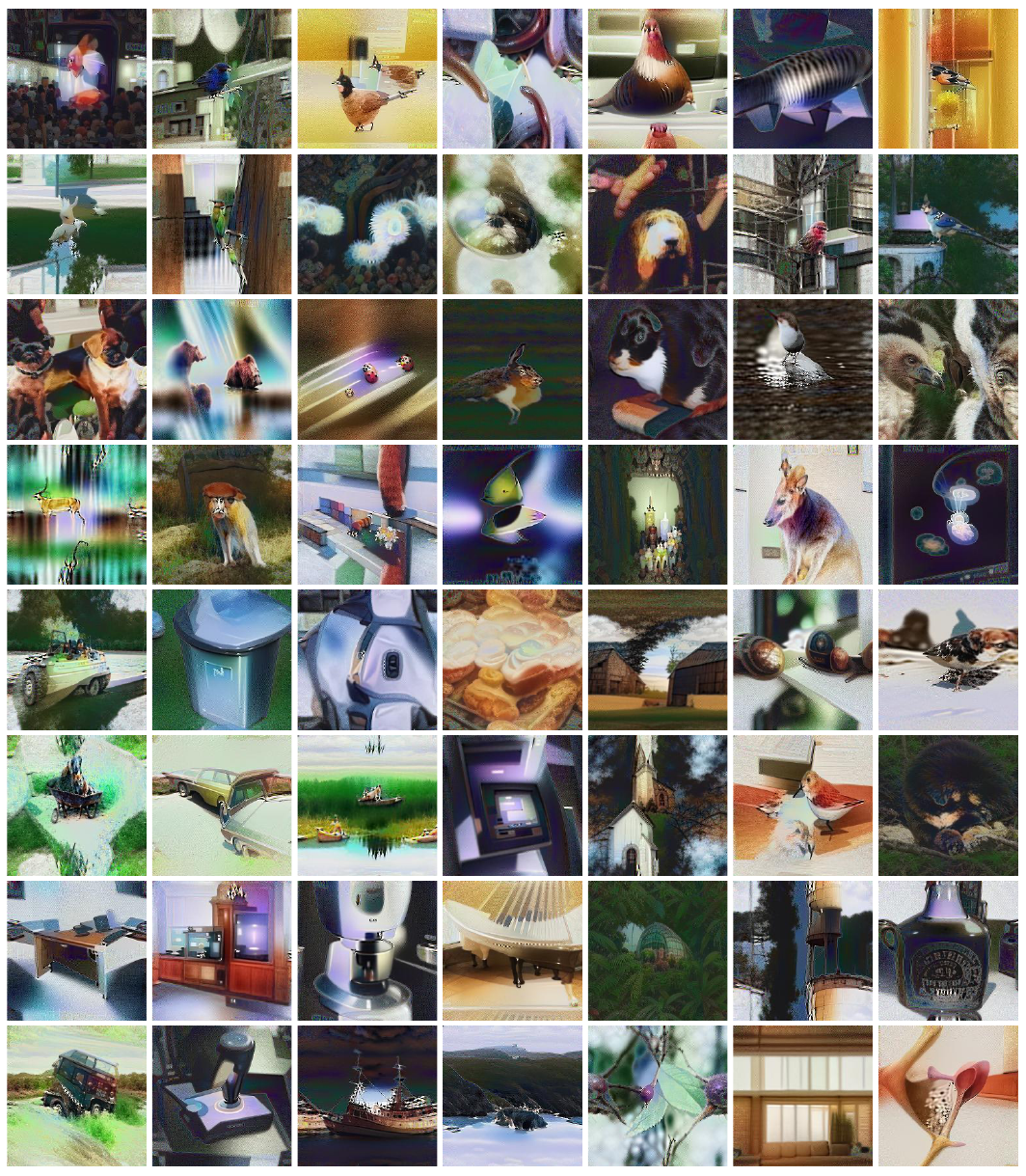}
		\vspace{-0.1in}
		\caption{Synthetic data on ImageNet-1K from our {\em SC-DD}.}
		\label{fig:IM_show}
	\end{figure*}

	Furthermore, we visualize more synthetic images derived from Tiny-ImageNet and ImageNet-1K datasets in Fig.~\ref{fig:tiny_contrast},~\ref{fig:IM_contrast} and~\ref{fig:IM_show}. We first provide the comparison of distilled images from MTT, SRe$^2$L, and our proposed approach on Tiny-ImageNet, as in Fig.~\ref{fig:tiny_contrast}. It can be observed that the images generated by MTT have richer textures and color information, but are less realistic for the object of the class. Then, we conduct a comparative visualization of the distilled images from our proposed method and SRe$^2$L on ImageNet-1K, as illustrated in Fig.~\ref{fig:IM_contrast}.
	
	By comparing images generated by various approaches, we illustrate that the images refined through our approach possess greater realism and encapsulate a broader spectrum of information than those produced by competing methods. This evidence demonstrates the enhanced effectiveness of the proposed approach.
	
	\clearpage
	
\end{document}